%% file: taslp.tex
\documentclass[lettersize,journal]{IEEEtran}
\usepackage{amsmath,amsfonts}
\usepackage{algorithm}
\usepackage{array}
\usepackage[caption=false,font=normalsize,labelfont=sf,textfont=sf]{subfig}
\usepackage{textcomp}
\usepackage{stfloats}
\usepackage{url}
\usepackage{verbatim}
\usepackage{graphicx}
\usepackage{cite}
\usepackage[utf8]{inputenc}
\usepackage{booktabs}
\usepackage{graphicx}
\usepackage{CJK}
\usepackage{multirow}
\usepackage{color}
\usepackage{verbatim}
\usepackage{url}
\usepackage{enumitem}
\usepackage{amsmath}
\usepackage{diagbox}
\usepackage{flushend,cuted}
\usepackage{bbm}
\usepackage{xspace}
\usepackage{algorithm}
\usepackage{algpseudocode}
\usepackage{colortbl}
\usepackage{bbding}
\usepackage{amssymb}
\DeclareMathOperator*{\argmax}{arg\,max}

\usepackage{listings}
\usepackage{ulem}
\usepackage{listings}
\usepackage{wrapfig}
\usepackage{soul}
\usepackage{makecell}

\usepackage{xcolor}

\newcommand{\kaiti}[1]{\begin{CJK*}{UTF8}{gkai} #1 \end{CJK*}}
\soulregister{\kaiti}7
\usepackage{pifont}
\usepackage{arydshln}
\usepackage{booktabs}
\usepackage{textcomp}
\newcommand{\mycircled}[1]{%
   \raisebox{2pt}{\textcircled{\raisebox{-0.9pt}{\kern-0.2pt #1}}}%
}
\usepackage{rotating}
\usepackage{textcomp}
\usepackage{stfloats}
\usepackage{amsmath}

\usepackage[hidelinks]{hyperref}

\begin{document}

\title{The Power of Question Alignment in Multilingual Reasoning: Broadened Scope and Deepened Insights}


\author{Wenhao Zhu, Shujian Huang, Fei Yuan, Cheng Chen, Jiajun Chen, Alexandra Birch
\thanks{Wenhao Zhu, Shujian Huang, Cheng Chen and Jiajun Chen are with National Key Laboratory for Novel Software Technology, Nanjing Univeristy, China. E-mail: \texttt{\{zhuwh,chengchen\}@smail.nju.edu.cn, \{huangsj, chenjj\}@nju.edu.cn}}
\thanks{Fei Yuan is with Shanghai Artificial Intelligence Laboratory, China. E-mail: \texttt{yuanfei@pjlab.org.cn}}
\thanks{Alexandra Birch is with School of Informatics, University of Edinburgh, United Kingdom. E-mail: \texttt{a.birch@ed.ac.uk}}}




\maketitle

\input{Latex/00_abstract}


\input{Latex/01_introduction}

\input{Latex/02_related_work}

\input{Latex/03_method}

\input{Latex/04_setting}

\input{Latex/05_experiments}

\input{Latex/06_analysis}

\input{Latex/07_conclusion}

\normalem
\bibliography{TASLP}
\bibliographystyle{IEEEtran}

\input{Latex/09_biography}

\end{document}

%% file: Latex/00_abstract.tex
\begin{abstract}
Bridging the significant gap between large language model's English and non-English performance presents a great challenge.
While some previous studies attempt to mitigate this gap with translated training data, the recently proposed question alignment framework leverages the model's English expertise to improve multilingual performance with minimum usage of expensive, error-prone translation.
In this paper, we explore how broadly this method can be applied by examining its effects in reasoning with and without chain-of-thought, as well as with program-of-thought.
We also explore applying this framework to extremely large language models in an efficient manner, such as through proxy-tuning.
Experiment results on multilingual reasoning benchmarks \textsc{mGSM}, \textsc{mSVAMP}, \textsc{xCSQA} and \textsc{xNLI} demonstrate that we can extend question alignment framework to boost multilingual performance across diverse reasoning scenarios, model families, and sizes.
For instance, when applied to the LLaMA2 models, it brings an average accuracy improvements of 12.2\% on \textsc{mGSM} even with the 70B model.
To understand the mechanism of its success, we analyze representation space, generated response and data scales, and reveal how question translation training strengthens language alignment within LLMs and shapes their working patterns.
\end{abstract}

%% file: Latex/01_introduction.tex
\section{Introduction}
Although large language models (LLMs) have shown the potential to solve complex reasoning problems~\cite{hendrycks2021measuring,kojima2022large,chowdhery2023palm,touvron2023llama}, LLMs still struggle in multilingual contexts~\cite{shi2022language,huang2023not,qin2023cross}.
This is unsurprising, given that their training data is predominantly made of English corpus and instructions~\cite{blevins2022language, wang2023far}. 
However, an important challenge remains: how to improve LLM performance on reasoning tasks in languages other than English with scarce multilingual resources. 
Previous studies attempt to collect multilingual data for continued pre-training~\cite{nguyen2023seallms} or instruction-tuning~\cite{chen2023breaking}.
However, training for reasoning across potentially dozens or hundreds of languages is both costly and inefficient, and sometimes the necessary resources are either unavailable or of poor quality~\cite{zhu2024question}.

\begin{figure}[t]
    \centering
    \includegraphics[width=0.45\textwidth]{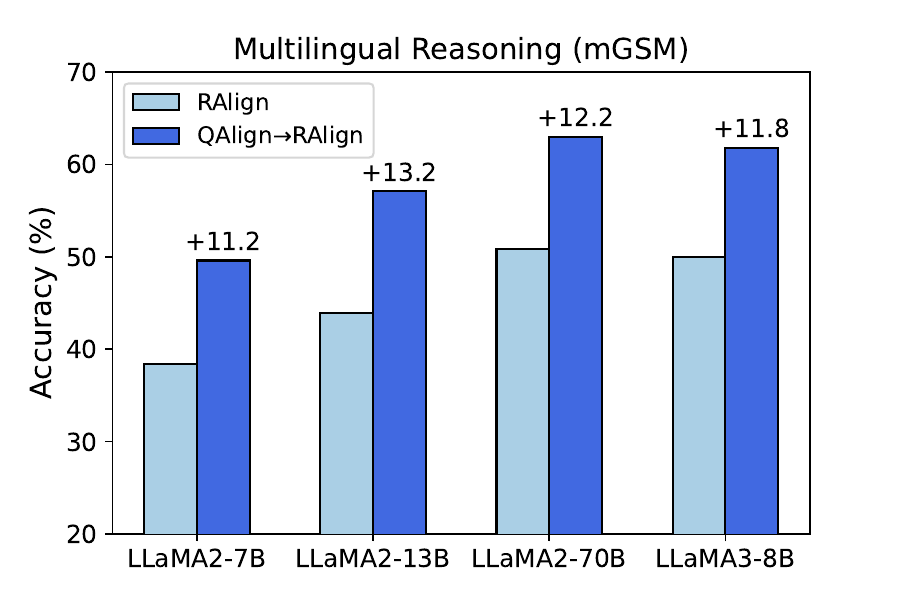}
    \caption{An illustration of the benefit of fine-tuning on question translation data (question alignment, QAlign) compared to standard fine-tuning on question-response instruction pairs (response alignment, RAlign). The added QAlign stage enhances the performance of LLaMA models across ten languages. Experiment results on more reasoning scenarios, model families and sizes will be reported in the experiment section.}
   \label{fig:intro}
\end{figure}

A compelling solution is to leverage the LLM's English proficiency in non-English contexts~\cite{shi2022language,sherborne2023optimal}.
Focusing on multilingual mathematical reasoning, Zhu et al. \cite{zhu2024question} demonstrate that their devised two-step training framework can effectively enable the sharing of English reasoning capabilities across different languages. 
Specifically, they employ question translation training on the pre-trained LLM to enhance its internal language alignment (referred to as question alignment) and then utilize cutting-edge English instruction data (question-response pairs) for response alignment to unlock LLM's reasoning capabilities in multilingual contexts. 

Despite the progress made in the previous work, there remain three limitations: (1) \textit{single reasoning scenario}: the scope of consideration was limited to math reasoning with chain-of-thought, with other reasoning scenarios not yet being taken into account. 
(2) \textit{opaque training effects}: the impact of the two-stage training on the LLM's working patterns and each training stage's contribution remain unclear.
(3) \textit{limited model size}: the effectiveness of this approach has yet to be confirmed on extremely large dense or sparse language models.

In this paper, we comprehensively address these left limitations.
First, we examine how broadly applicable the training framework is across varied reasoning scenarios. 
These involve distinict types of problem-solving languages and reasoning objectives: math reasoning with chain-of-thought~\cite{wei2022chain}, math reasoning with program-of-thought~\cite{chen2023program}, common sense reasoning without intermediate thought~\cite{conneau2018xnli,talmor2019commonsenseqa}.
We demonstrate that the instruction-tuned model consistently benefits from the added question alignment stage in different reasoning scenarios.

\begin{figure*}
    \centering
    \includegraphics[width=0.8\textwidth]{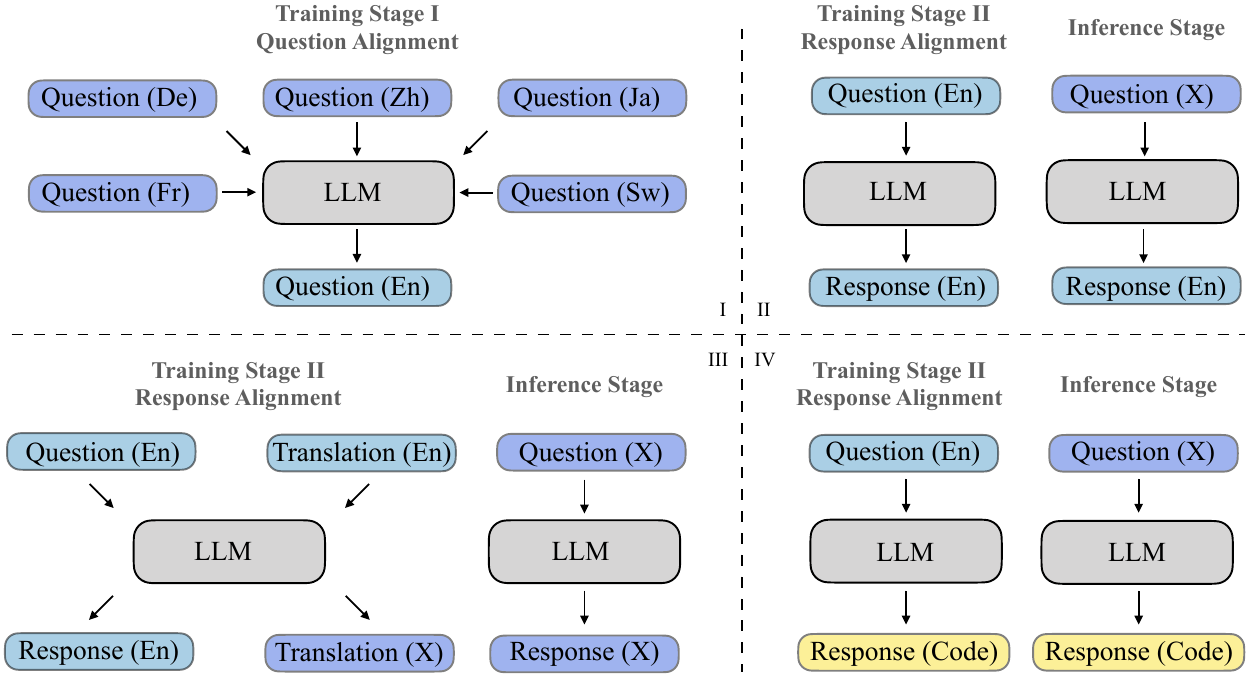}
   \caption{Illustration of the original two-step training framework~\cite{zhu2024question} (shown in Subfigures I and II) and our extension (Subfigures III and IV and described in Section~\ref{sec:extend}). Subfigure I and II illustrate the training and inference process of the orignal training framework. In subfigure III and IV, by maintaining the question alignment stage unchanged and modifying the response alignment stage, we adapt this framework to a wider range of scenarios. For example, in Subfigure IV, we use code instruction data for the second stage of training to unlock the LLM's capability for reasoning with program-of-thought. In subfigure III, we incorporate En-X translation data in the second stage of training and attempt to bias the LLM to generate non-English response.}
   \label{fig:illustration}
\end{figure*}

By keeping the question alignment stage the same and altering the response alignment stage, we adapt this framework for a broader range of scenarios. For instance, in Subfigure IV (Fig. 2), we use code instruction data in the second stage to enhance the LLM's reasoning capabilities with program-of-thought. In Subfigure III (Fig. 2), we incorporate En-X translation data in the second stage to encourage the LLM to generate responses in languages other than English.

To gain clearer insights, we analyze how our instruction-tuned model utilizes its English reasoning capability in non-English contexts.
Our analysis begins with the representation space, which fundamentally determines how the model derives the output answer.
We discover that question translation training significantly affects the distribution of multilingual representations in LLM's middle to top layers, aligning them within the same semantic space as English (Figure~\ref{fig:representation}).
Additionally, we examine the LLM’s generated response step-by-step and find that the unified representation space makes the model to employ more similar problem-solving process to address multilingual questions.

Besides, we investigate how well question alignment aproach scales and whether it offers benefits to the most powerful open-source LLMs.
We explore applying question alignment to extremely large language models, e.g., LLaMA2-70B~\cite{touvron2023llama}.
While fully fine-tuning LLM's parameters is straightforward, the computation cost becomes prohibitive as the number of parameters increases significantly.
In constrast, we illustrate that proxy-tuning~\cite{liu2024tuning} offers as an efficient alternative for both dense models and sparse Mixture-of-Experts (MoE) models.
We also discover that by carefully selecting the proxy model, we can achieve 98.1\% performance of fully fine-tuning without updating any parameters.

Experiments on several multilingual reasoning benchmarks \textsc{mGSM}, \textsc{mSVAMP}, \textsc{xCSQA} and \textsc{xNLI} demonstrate that the question alignment approach is a broad technique applicable across a wide variety of difficult reasoning tasks and the modulaized training pipeline enables us to efficiently unlocks LLM's multilingual capabilities in the targeted skills.
By applying question alignment to extremely large language models, we find that it scales well and further strengthen those powerful LLM's multilingual reasoning abilities.
For instance, our instruction-tuned LLaMA2-70B achieves 63.0\% average accuracy on \textsc{mGSM} (Fig.~\ref{fig:intro}), pushing the multilingual reasoning performance of open-source models to a new height.

%% file: Latex/02_related_work.tex
\section{Related Work}
Extensive empirical analyses have shown that LLMs face challenges in multilingual contexts, especially with low-resource languages~\cite{shi2022language, huang2023not,zhu2023multilingual}.
In this paper, we focus on the core capability of LLM, i.e., the reasoning capability~\cite{meta2024llama3,reid2024gemini}, and explore to push forward the boundaries of LLMs in multilingual reasoning.

\subsection{Advancing LLM's Multilingual Capability}
Given that English predominates in the pretraining data~\cite{blevins2022language,touvron2023llama}, it is unsurprising that the model exhibits unbalanced multilingual performance.
To improve the multilingual capabilities of open-source LLMs, Nguyen et al.~\cite{nguyen2023seallms}, Lu et al.~\cite{lu2024llamax}, Fujii et al.~\cite{fujii2024continual} and Dou et al.~\cite{dou2024sailor} perform continue-pretraining with a massive non-English corpus, but this approach requires huge computational resources and is not data-efficient.
Chen et al.~\cite{chen2023breaking}, She et al.~\cite{she2024mapo} follow the translate-training paradigm~\cite{artetxe2020translation} and use machine-translated instruction data for fine-tuning.
Although multilingual instruction-tuning yields moderate improvements, continuously translating an increasing volume of English instruction data is usually prohibitive and translating lengthy, logical response can sometimes be error-prone~\cite{zhu2024question}.
Therefore, instead of relying on collecting more multilingual data, we explore a more effcient paradigm by enabling LLMs to leverage English expertise in non-English scenarios.

\subsection{Transferring LLM's English Expertise}
To enable LLMs to leverage its English expertise in non-English scenarios, the research community explores both explicit and implicit solutions.
The basic explicit method is to prompt the LLM to use an English chain-of-thought to solve non-English questions~\cite{shi2022language}.
Shi et al.~\cite{shi2022language} discover that intermediate reasoning steps in English consistently lead to competitive or better results than those written in the native language of the question.
Besides, Shi et al.~\cite{shi2022language}, Huang et al.~\cite{huang2023not} and Qin et al.~\cite{qin2023cross} explore to prompt LLM to translate non-English questions into English, then generate responses based on these translations.
However, the explict translation process increases inference cost and this approach is not always effective for LLMs lacking robust multilingual translation capabilities~\cite{zhu2024question}.
Recently, Zhu et al.~\cite{zhu2024question} propose a two-step instruction-tuning framework to guide LLMs to implicitly relate non-English questions to their English counterparts.
This is achieved through question translation training, allowing the LLM to utilize its English expertise to solve reasoning problems during inference.
In line with this philosophy, Yoon et al.~\cite{yoon2024langbridge} also recognize the importance of building alignment across multilingual questions and explore the use of an additional multilingual encoder, the encoder of \textsc{mT5}~\cite{xue2021mt5}, to map multilingual questions into the LLM's English semantic space for unified processing.
In this paper, we build upon the pioneering work of Zhu et al.~\cite{zhu2024question} and delve deeper to broaden the scope of that work and provide deepened insights.

\subsection{Modularized Training vs. Multi-task Training}
Unlike tranditional multi-task trainng~\cite{muennighoff2023crosslingual,wang2023far,singh2024aya}, the two-stage training framework we explore follows a modularized training fashion~\cite{pfeiffer2023modular}, which disentangles the training process into question translation training and question-response supervised training.
Zhu et al.~\cite{zhu2024question} demonstrated that training in the correct order is crucial for successfully transferring LLM's English expertise, while combining both stages into a single multi-task training is ineffective. 
In this paper, we further show how the added translation training stage shapes LLM's working pattern and boost its multilingual performance in various reasoning scenarios.

%% file: Latex/03_method.tex
\section{Methodology}
In this section, we will first recap the two-step training framework proposed by~\cite{zhu2024question} (Section \ref{sec:recap}).
Next, we introduce the diverse reasoning scenarios to examine the broad applicability of the training framework (Section \ref{sec:extend}).
Then, we present the flexible implementation of the question alignment stage and discuss its benefits (Section \ref{sec:flexible}).
Finally, we introduce efficient recipes to scale this framework to extremely large language models (Section \ref{sec:tuning}).

\subsection{Recap: Two-stage Training Framework}
\label{sec:recap}
The original training framework consists of two stages: \textit{question alignment} and \textit{response alignment}.
During quesiton alignment, X-En question translation data $(\mathcal{X}_{e}, \mathcal{X}_l)$ from $\mathcal{D}_l$ is utilized to train the model to associate non-English questions $\mathcal{X}_l$ with their English counterparts $\mathcal{X}_{e}$ (Figure~\ref{fig:illustration}, Subfigure I).
This process enhances language alignment within the large language model.
The optimization objective can be written as:
\begin{equation}
\argmax_{\theta} \ \sum_{l\in\mathcal{L}}\sum_{\{\mathcal{X}_e,\mathcal{X}_l\}\in\mathcal{D}_l} \log p_{\theta}(\mathcal{X}_{e} | \mathcal{X}_l) \nonumber
\end{equation}
where $l$ is the target language, $\mathcal{L}$ is the set of considered non-English languages, and $\theta$ denotes the model's parameters.

During response alignment, English cutting-edge instruction data $\mathcal{D}_e$ is used to unlock LLM's English reasoning capability (Figure~\ref{fig:illustration}, Subfigure II).
The optimization objective can be written as:
\begin{equation}
\argmax_{\theta} \sum_{\{\mathcal{X}_e,\mathcal{Y}_e\}\in\mathcal{D}_e}\log p_{\theta}(\mathcal{Y}_e | \mathcal{X}_e) \nonumber
\end{equation}
where $\mathcal{X}_e$ and $\mathcal{Y}_e$ denotes the English question and its corresponding response respectively.
During inference, thanks to the previously established language alignment, the LLM can effectively leverage its English expertise in non-English question contexts. 
Compared to the explicit translate-test approach, this framework is more efficient during inference and achieves higher reasoning accuracy by avoiding potential error accumulation~\cite{zhu2024question}.

\subsection{Explored Diverse Reasoning Scenarios}
\label{sec:extend}
Next, we introduce the diverse reasoning scenarios and illustrate how we adapt the original training framework to these scenarios (Figure \ref{fig:illustration}).

\subsubsection{Math Reasoning with chain-of-thought}
\begin{figure}[htbp]
    \centering
    \includegraphics[width=0.45\textwidth]{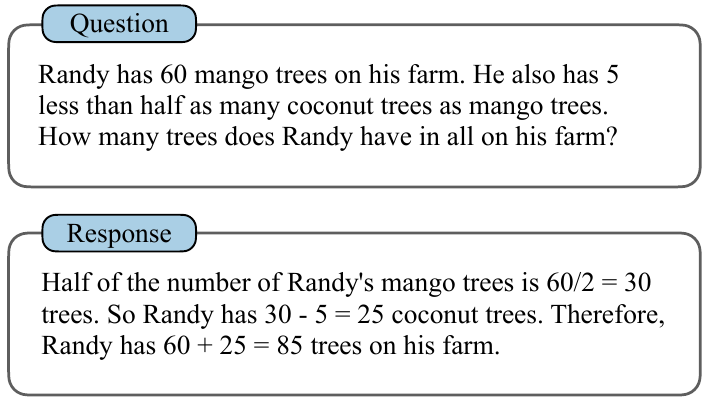}
   \caption{Illustration of the employed instruction data. We use this instruction data to teach model to solve mathematical reasoning task with chain-of-thought.}
   \label{fig:illustration1}
\end{figure}

The first reasoning scenario is solving mathematical reasoning task through chain-of-thought (CoT).
To unlock the model's capability on this, we follow Zhu et al.~\cite{zhu2024question} and utilize the English instruction data depicted in Figure~\ref{fig:illustration1}.
We observe that when fine-tuned in this manner, the model tends to solve multilingual questions using English chain-of-thought.
Previous findings indicate that this is beneficial, as LLMs typically achieve higher reasoning accuracy with English CoT than with non-English CoT~\cite{shi2022language}.
However, this may be unsatisfactory for users from diverse linguistic backgrounds who prefer responses in their query language, which help them better understand the model’s problem-solving process~\cite{kew2023turning}.

To support multilingual chain-of-thought generation, we discover that mixing English instruction data with En-X general domain translation data during the response alignment stage helps (Figure~\ref{fig:illustration}, Subfigure III).
We observe that performing multi-task training in training stage II enables the model to naturally produce multilingual CoTs in response to multilingual queries, even when no multilingual question-response pairs are available during instruction-tuning.
This training strategy aligns with prior research that incorporates translation tasks into multi-task fine-tuning~\cite{zhu2023extrapolating,zhang2023bayling}, but our work investigated it in a more controlled setting and explores the impact of multi-task training on the output language of CoT.

\subsubsection{Math reasoning with program-of-thought}
\begin{figure}[ht]
    \centering
    \includegraphics[width=0.45\textwidth]{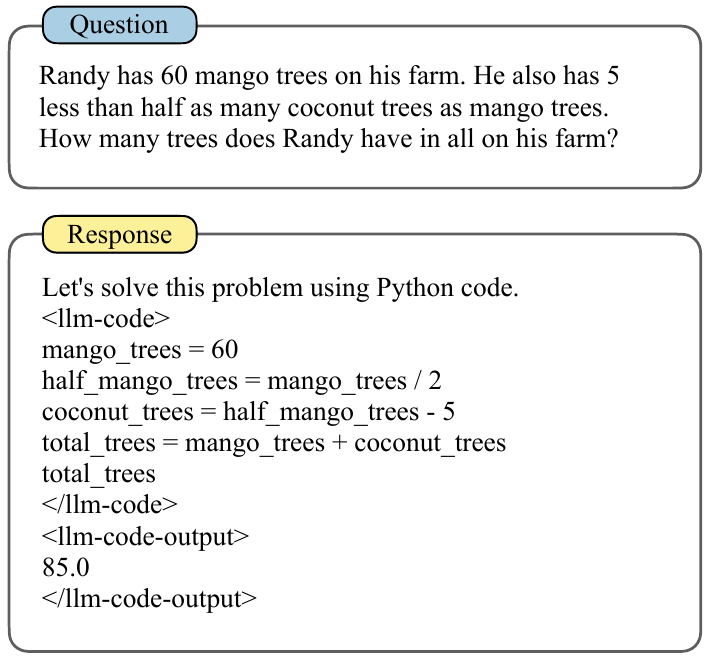}
   \caption{Illustration of the employed instruction data. We use this instruction data to teach model to solve mathematical reasoning task with program-of-thought.}
   \label{fig:illustration2}
\end{figure}

The second reasoning scenario involves solving mathematical reasoning task with executable code~\cite{gao2023pal,chen2023program,shi2023large}.
In this scenario, the response consists of Python code instead of a natural language chain-of-thought.
A major advantage of reasoning through executable code is that it helps avoid basic computational errors~\cite{chen2023program}.
To unlock the model's capability on this, we utilize the instruction data depicted in Figure~\ref{fig:illustration2}.
Specifically, the model needs to generate Python code enclosed within ``\textless llm-code\textgreater'' and ``\textless/llm-code\textgreater'' tags.
Then a Python interpreter will be used to execute this code block.
The derived numerical answer, corresponding to the value of the variable in the last line of the code block, will be enclosed within ``\textless llm-code-output\textgreater'' and ``\textless/llm-code-output\textgreater'' tags.

\subsubsection{Common sense reasoning without intermediate thought}
\begin{figure}[htbp]
    \centering
    \includegraphics[width=0.45\textwidth]{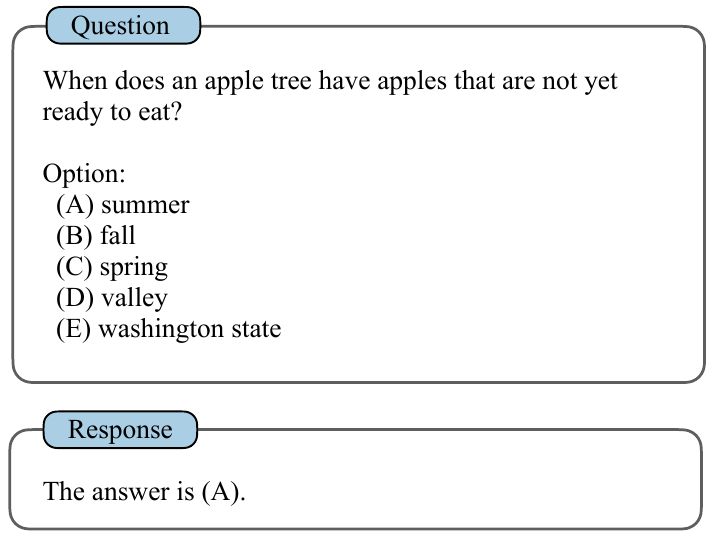}
   \caption{Illustration of the employed instruction data. We use this instruction data to teach model to solve common sense reasoning task.}
   \label{fig:illustration3}
\end{figure}

The third reasoning scenario is common sense reasoning.
Unlike math reasoning, this scenario requires the model to answer the given question using the common sense knowledge embedded in its parameters.
This scenario does not involve a chain-of-thought process.
To unlock the model's capability on this, we utilize the instruction data depicted in Figure~\ref{fig:illustration3}.
In this task, the model needs to select the appropriate answer from the given options through common sense reasoning.

\subsection{Flexible and Combinable Implementation of Question Alignment Stage} 
\label{sec:flexible}
Next, we present the flexible implementation of the question alignment stage.
While the original implementation relies on question translation data, we discover that similar goal can be achieved by using general domain X-En translation data, which allows a more flexible implementation of the question alignment stage. 
This discovery facilitates a more adaptable and versatile execution of the question alignment process.

More importantly, by combining question translation data $\mathcal{D}_l$ with general domain translation data $\mathcal{\tilde{D}}_l$, we can leverage the value of both data and further enhance the model's understanding on non-English questions.
\begin{equation}
\argmax_{\theta} \ \sum_{l\in\mathcal{L}}\sum_{\{\mathcal{X}_e,\mathcal{X}_l\}\in\mathcal{D}_l\cup\mathcal{\tilde{D}}_l} \log p_{\theta}(\mathcal{X}_{e} | \mathcal{X}_l) \nonumber
\end{equation}

\subsection{Scaling to Extremely Large LM}
\label{sec:tuning}
Extremely large language models, e.g., LLaMA2-70B, Mixtral-8$\times$22B, often demonstrate state-of-the-art performance among open-source LLMs.
We are eager to see whether we can apply this framework effectively and efficiently to those scaled-up models and push the multilingual reasoning performance of open-source models to a new height.

\subsubsection{Vanilla fine-tuning} While fully fine-tuning extremely large models is a straightforward solution, the process can be prohibitively time-consuming and computationally expensive, due to the vast number of parameters that need to be optimized.

\subsubsection{Efficient proxy-tuning} Inspired by Liu et al.~\cite{liu2024tuning}, we explore proxy-tuning as an efficient alternative solution. 
The objective of proxy-tuning is to guide a large pre-trained model $\mathcal{M}$ to behave like a tuned model without updating any parameters.
This is achieved by employing a small pre-trained model $\mathcal{M}^-$ and a small tuned model $\mathcal{M}^+$ to serve as the expert model and the anti-expert model\footnote{The small models must have the same vocabulary as the large model to support arithmetic operations on prediction distributions. }.
The underlying assumption of proxy-tuning is that the difference in logits between $\mathcal{M}^-$ and $\mathcal{M}^+$ can approximate the difference between $\mathcal{M}$ and a truly-tuned large model.
At each inference step $t$, we condition the base model $\mathcal{M}$, the expert $\mathcal{M}^+$ and the anti-expert $\mathcal{M}^-$ on the question $\mathcal{X}$ and the generated response prefix $\mathcal{Y}_{<t}$. 
The probability distribution for the proxy-tuned model $\hat{\mathcal{M}}$ is derived from the prediction distributions of these models:

{\normalsize
\begin{equation}
p_{\hat{\mathcal{M}}}(\mathcal{Y}_t|\mathcal{X}, \mathcal{Y}_{<t}) \propto  p_{\mathcal{M}}(\mathcal{Y}_t|\mathcal{X}, \mathcal{Y}_{<t})\frac{p_{\mathcal{M}^+}(\mathcal{Y}_t|\mathcal{X}, \mathcal{Y}_{<t})}{p_{\mathcal{M}^-}(\mathcal{Y}_t|\mathcal{X}, \mathcal{Y}_{<t})} \nonumber 
\label{eq:proxy}
\end{equation}
}

\begingroup
\renewcommand{\arraystretch}{1.2} 
\begin{table*}[htbp]
\centering
\footnotesize
\caption{Statistics of involved datasets. ``\#Lang'' denotes the number of languages covered by the dataset and ``\#Sample'' refers to the total number of samples it contains. The labels ``Question'' and ``Response'' denotes whether each sample includes a question and a corresponding response. The symbols \mycircled{1}\mycircled{2}\mycircled{3} correspond to the three scenarios discussed in Section~\ref{sec:extend}: reasoning with chain-of-thought, reasoning with program-of-thought and reasoning withouth intermediate thought.}
\begin{tabular}{p{3cm}p{1.1cm}<{\centering}p{1.4cm}<{\centering}p{1.2cm}<{\centering}p{1.2cm}<{\centering}p{1.5cm}<{\centering}p{1.2cm}<{\centering}}
\toprule
\hspace{1cm}\textbf{Dataset}                   & \textbf{\# Lang} & \textbf{\# Sample} & \textbf{Usage} & \textbf{Question} & \textbf{Response} & \textbf{Scenario} \\
\midrule
\textsc{MetaMathQA}       & 1     & 395,000   & Training & \Checkmark & \Checkmark & \mycircled{1} \\
\textsc{OpenMathInstruct} & 1     & 1,343,849 & Training & \Checkmark & \Checkmark & \mycircled{2} \\ 
\textsc{GSM8KInstruct}    & 10    & 73,559    & Training & \Checkmark & \XSolidBrush  & \mycircled{1}\mycircled{2} \\
\textsc{mGSM}             & 10    & 2,500    & Evaluation & \Checkmark & \Checkmark & \mycircled{1}\mycircled{2} \\
\textsc{mSVAMP}           & 10    & 10,000   & Evaluation & \Checkmark & \Checkmark  & \mycircled{1}\mycircled{2} \\
\hdashline
\textsc{XCSQA-train}     & 1      & 8,888    & Training & \Checkmark & \Checkmark & \mycircled{3}  \\
\textsc{XCSQA-test}      & 15     & 17,184   & Training & \Checkmark & \XSolidBrush & \mycircled{3} \\
\textsc{XCSQA-dev}       & 15     & 16,000   & Evaluation & \Checkmark & \Checkmark  & \mycircled{3} \\
\hdashline
\textsc{MultiNLI}        & 1      & 392,702  & Training & \Checkmark & \Checkmark & \mycircled{3}  \\
\textsc{XNLI-dev}        & 15     & 34,860  & Training & \Checkmark & \XSolidBrush & \mycircled{3} \\
\textsc{XNLI-test}       & 15     & 75,150 & Evaluation & \Checkmark & \Checkmark  & \mycircled{3} \\
\bottomrule
\end{tabular}
\label{tab:statistics}
\end{table*}
\endgroup

Moreover, we discover that the selection of the small expert and anti-expert model is crucial for the final performance.
By carefully selecting small models for proxy-tuning, such as using a 13B model instead of a 7B model to proxy-tune the 70B model, we can almost achieve the performance of fully fine-tuning without updating any parameters.

%% file: Latex/04_setting.tex
\section{Experiment setting}
\subsection{Base Models} 
We consider a range of the most powerful open-source pre-trained LLMs for our experiments.
In most experiments, we use LLaMA2-7B as the base model.
In experiments involving reasoning with program-of-thought, we use CodeLLaMA-7B~\cite{roziere2023code} as the base model.
In experiments involving extremely large language models, we consider LLaMA2-70B, LLaMA3-70B~\cite{meta2024llama3}, Mixtral-8x7B~\cite{jiang2024mixtral} and Mixtral-8x22B~\cite{mistral2024mixtral}.

\subsection{Training dataset}
In mathematical reasoning, we use multilingual questions from \textsc{GSM8KInstruct}~\cite{chen2023breaking} for question alignment. 
During response alignment, we use \textsc{MetaMathQA}~\cite{yuan2023scaling} to teach LLM to reason with chain-of-thought.
To teach LLM to reason with program-of-thought, we use \textsc{OpenMathInstruct}~\cite{toshniwal2024openmathinstruct}.
In common sense reasoning, we use multilingual questions from \textsc{XCSQA-test}~\cite{lin2021xcsr} and \textsc{xNLI-dev}~\cite{conneau2018xnli} for question alignment and use English supervised data in \textsc{XCSQA-train}~\cite{lin2021xcsr} and \textsc{MultiNLI}~\cite{williams2018broad} for response alignment.
Statistics of involved datasets are reported in Table~\ref{tab:statistics}.
Besides, we illustrate the flexible and combinable implementation of the question alignment stage by using multilingual translation data from \textsc{WikiMatrix}~\cite{schwenk2021wikimatrix} dataset.

\subsection{Training details}
Following Zhu et al.~\cite{zhu2024question}, we use \textit{QAlign}\footnote{\url{https://github.com/NJUNLP/QAlign}} as the code base.
We use consistent training hyper-parameters across two stages of training.
At each stage, we fine-tune LLM's full parameters for 3 epochs on 8$\times$A100 GPUs.
The learning rate is 2e-5, with a batch size of 128.

\subsection{Baseline Methods}
The direct baseline for the two-step training framework is solely performing response alignment (RAlign), i.e. fine-tuning with monolingual reasoning data.
Comparing this baseline with the alignment-enhanced model (QAlign$\rightarrow$RAlign) directly illustrates the benefits of performing question alignment and the effectiveness of leveraging English expertise.
Additionally, we compare our approach with other solutions, including continued pre-training and specialized instruction-tuning, to comprehensively illustrate the priority of the question alignment framework.

\begin{figure*}[!ht]
    \centering
    \includegraphics[width=0.95\textwidth]{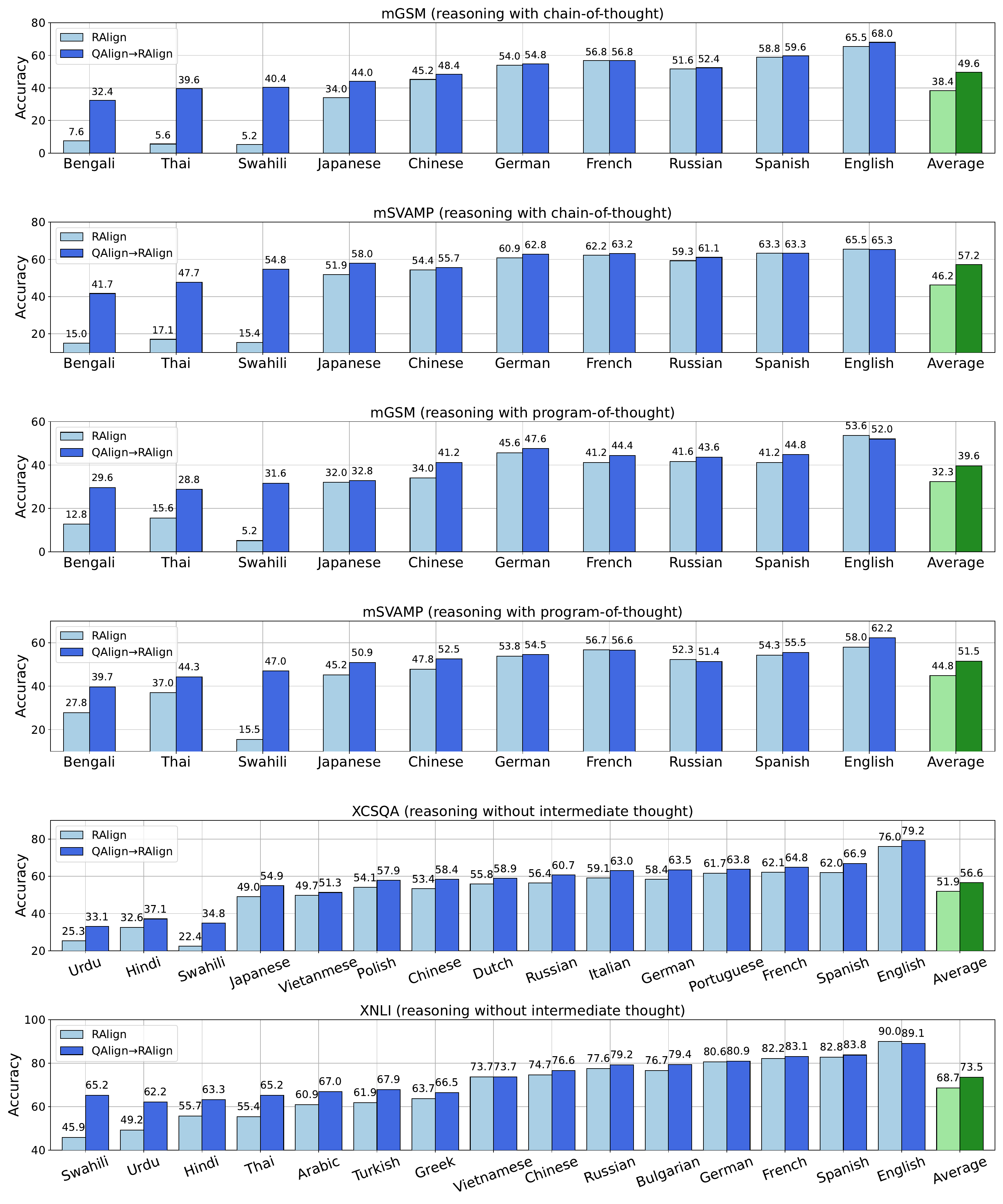}
    \caption{Reasoning accuracy on \textsc{mGSM}, \textsc{mSVAMP}, \textsc{xCSQA} and \textsc{xNLI}. The improvement in multilingual performance is evident in all three reasoning scenarios: reasoning with chain-of-thought, reasoning with program-of-thought and reasoning without intermediate thought.}
    \label{fig:diverse}
\end{figure*}

\subsection{Evaluation dataset}
We use \textsc{mGSM}~\cite{shi2022language} and \textsc{mSVAMP}~\cite{chen2023breaking} to evaluate LLM's performance on multilingual mathematical reasoning. 
We use \textsc{XCSQA-dev}~\cite{talmor2019commonsenseqa} and \textsc{xNLI-test}~\cite{conneau2018xnli} to evaluate LLM's performance on multilingual common sense reasoning.
Dataset statistics are reported in Table~\ref{tab:statistics}.
We report LLM's answer accuracy in a zero-shot and greey-decoding setting.
We measure answer accuracy by comparing the last numerical value (for \textsc{mGSM}, \textsc{mSVAMP}) / the label within brackets (for \textsc{xCSQA}) / the output label (for \textsc{xNLI}) that appears in the LLM-generated response with the gold answer.

%% file: Latex/05_experiments.tex
\section{Experiment results}
In this section, we will report experiment results and introduce our main findings.

\begin{figure}[!ht]
    \centering
    \includegraphics[width=0.45\textwidth]{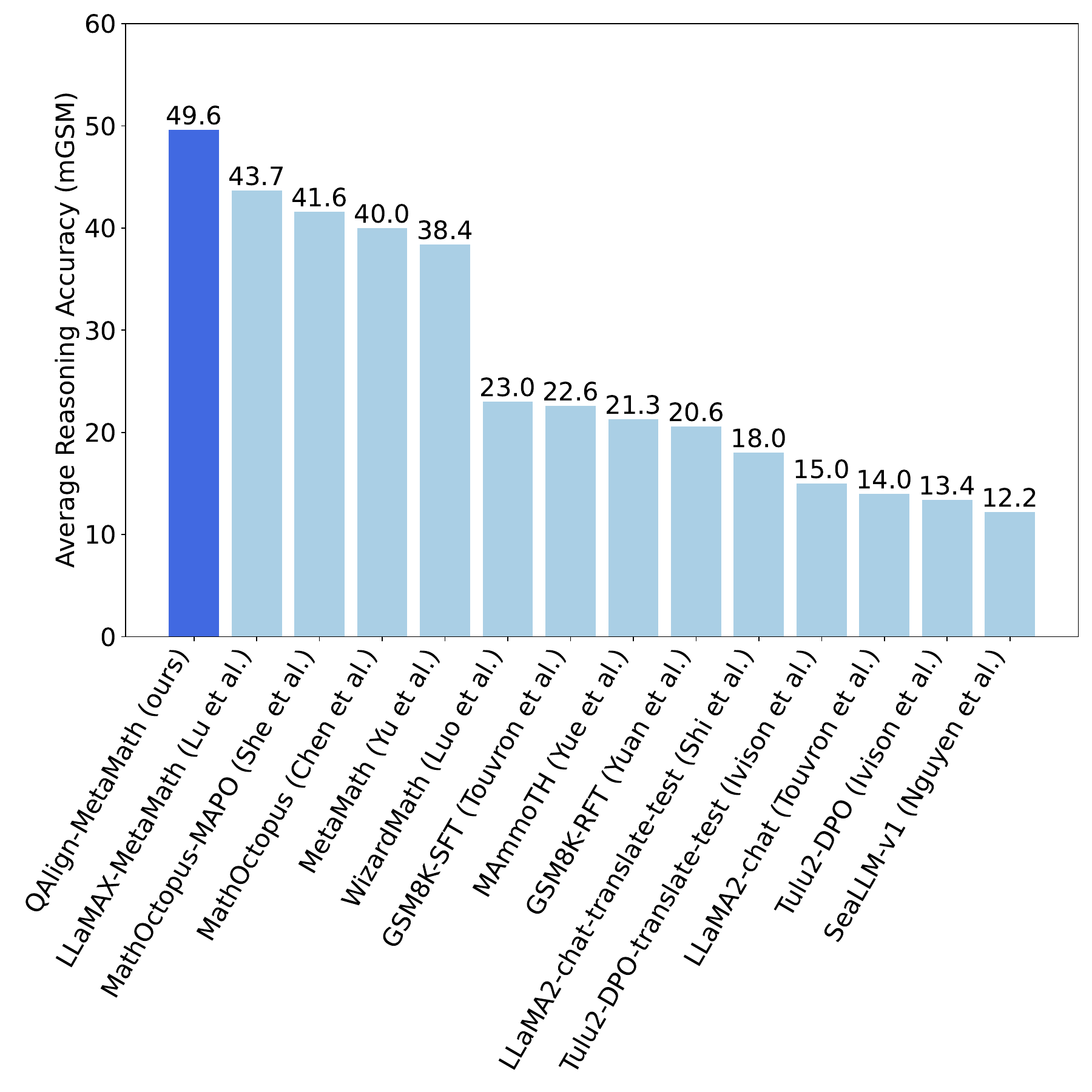}
    \caption{Comparison across different approaches on \textsc{mGSM} dataset. For fair comparison, all models we consider are developed from LLaMA2-7B.}
    \label{fig:benchmark}
\end{figure}

\begingroup
\renewcommand{\arraystretch}{1.4} 
\begin{table*}[ht]
\centering
\small
\caption{Effects of incorporating En-X general domain translation data for response alignment training. En-X translation training implicitly bias LLM to generate non-English chain-of-thought and increase the question-response language consistency.}
\begin{tabular}{lccccccccccc}
\hline
\multirow{2}{*}{\hspace{1.2cm}\textbf{System (7B)}} & \multicolumn{11}{c}{\textbf{Question-Response Language Consistency (\textsc{mGSM})}} \\
\cline{2-12}
 & \textbf{Bn} & \textbf{Th} & \textbf{Sw} & \textbf{Ja} & \textbf{Zh} & \textbf{De} & \textbf{Fr} & \textbf{Ru} & \textbf{Es} & \textbf{En} & \textbf{Avg.} \\
QAlign$\rightarrow$RAlign (\textit{w.o. En-X training}) & 0.0 & 0.0 & 2.9 & 0.0 & 0.0 & 0.5 & 0.1 & 0.1 & 0.0 & 93.8 & 9.7 \\
QAlign$\rightarrow$RAlign (\textit{w. En-X training})   & 26.8 & 42.7 & 49.3 & 63.1 & 26.8 & 63.2 & 36.9 & 82.4 & 37.9 & 93.4 & 52.3 \\
\hline
\multirow{2}{*}{\hspace{1.2cm}\textbf{System (7B)}} & \multicolumn{11}{c}{\textbf{Reasoning Accuracy (\textsc{mGSM})}} \\
\cline{2-12}
 & \textbf{Bn} & \textbf{Th} & \textbf{Sw} & \textbf{Ja} & \textbf{Zh} & \textbf{De} & \textbf{Fr} & \textbf{Ru} & \textbf{Es} & \textbf{En} & \textbf{Avg.} \\
QAlign$\rightarrow$RAlign (\textit{w.o. En-X translation}) & 32.4 & 39.6 & 40.4 & 44.0 & 48.4 & 54.8 & 56.8 & 52.4 & 59.6 & 68.0 & 49.6 \\
QAlign$\rightarrow$RAlign (\textit{w. En-X translation}) & 21.2 & 21.6 & 24.8 & 28.0 & 41.6 & 50.8 & 55.2 & 45.6 & 54.4 & 66.4 & 41.0 \\
\hline
\end{tabular}
\label{tab:language}
\end{table*}
\endgroup
\subsection{The question alignment framework is broadly applicable across diverse reasoning scenarios}
At first, we examine the trainining framework in various reasoning scenarios.
Experiment results on \textsc{mGSM}, \textsc{mSVAMP}, \textsc{XCSQA}, \textsc{xNLI} are depicted in Figure~\ref{fig:diverse}.
The improvement in multilingual performance is evident in all these reasoning scenarios: reasoning with chain-of-thought, with program-of-thought, and without intermediate thought.
These results demonstrate that performing question alignment before task-specific instruction-tuning can be broadly beneficial across diverse reasoning scenarios.
In Figure~\ref{fig:benchmark}, we also take math reasoning as a representative scenario and demonstrate that question aligment framework enjoys superioty over other previously proposed solutions.
\begin{table}[!htbp]
\centering
\small
\caption{Experiment results for different implementation of the question alignment stage. the first row denotes the model that does not have the question alignment stage. ``\#Pair'' refers to the total number of translation pairs we use. ``Lrl'' and ``Hrl'' denotes the average multilingual performance on non-English low-resource and high-resource languages respectively. The underlined text denotes the highest performance along the column.}
\begin{tabular}{p{2.6cm}p{1.2cm}p{0.5cm}<{\centering}p{0.5cm}<{\centering}p{0.5cm}<{\centering}}
\toprule
\multirow{2}{*}{\textbf{Translation Data}} & \multirow{2}{*}{\textbf{\#Pair}} & \multicolumn{3}{c}{\textsc{\textbf{mGSM (CoT)}}}   \\
                                  &                         & \textbf{Lrl} & \textbf{Hrl} & \textbf{En}  \\
\midrule
-                                    & -           & 6.4  & 50.1 & 65.6 \\
WikiMatrix                           & 64k         & 13.4 & 51.1 & 68.0 \\
WikiMatrix                           & 400k        & 23.4 & 51.0 & 67.2 \\
Question                             & 73k         & 36.4 & 52.7 & 68.0 \\
Question+WikiMatrix                  & 73k+400k    & \underline{39.6} & \underline{53.6} & 67.2 \\
\bottomrule
\end{tabular}
\label{tab:flexible}
\end{table}
We can see that the question alignment approach is not only more effective and data-efficient than continued pre-training approach, such as LLaMAX~\cite{lu2024llamax}, SeaLLM~\cite{nguyen2023seallms}, but it also surpasses other instruction-tuning strategies that do not specifically strengthen language alignment, such as MetaMath~\cite{yu2023metamath}, MathOctopus~\cite{zhao2021neurst}, WizardMath~\cite{luo2023wizardmath}, MAmmoTH~\cite{yue2023mammoth}, Tulu2~\cite{ivison2023camels} and LLaMA2-chat~\cite{touvron2023llama}.

\subsection{Incorporating En-X translation training during response alignment can bias LLM to generate non-English response}
As shown in Table~\ref{tab:language}, the original QAlign$\rightarrow$RAlign model often exhibits extremely low question-response language consistency\footnote{We perform language identification with the \textsc{OpenLID} toolkit~\cite{burchell2023open}.} for non-English tasks, as it tends to respond in English to multilingual questions.
After incorporating En-X translation training for stage II training, the fine-tuned model responds more frequently in the same language as the question, demonstrating a significant increase in language consistency (52.3\% vs. 9.7\%).
However, while this approach enhances language consistency, it also reduces reasoning accuracy.
This observation aligns with the findings of Shi et al.~\cite{shi2022language}, which suggest that current LLMs perform better with English CoT than with non-English CoT.
Future research needs to address this trade-off and find a better balance between language consistency and reasoning accuracy.
In the following experiments and analysis, for the sake of reasoning accuracy, we do not include translation data in stage II training to force LLM to generate non-English response.

\subsection{Combining question translation and general domain translation for question alignment stage could bring further improvements}
Next, we demonstrate the flexible and combinable implementation of the question alignment stage using general domain translation data\footnote{For each X-En directions, we sample at most 50k translation data from \textsc{WikiMatrix}. Since it does not cover Thai, we make observations on the remaining eight non-English languages and English. Following the definition in Shi et al.~\cite{shi2022language}, we divide these non-English languages into two groups in Table~\ref{tab:flexible}: low-resource langauges (Bn, Sw) and high-resource languages (Ja, Zh, De, Fr, Ru, Es).}.
Experiment results in Table \ref{tab:flexible} show that using general domain translation for question alignment stage can achieve a similar enhancement purpose, although it is less data-efficient compared to using in-domain question translation data.
The benefit of discovering this alternative implementation is that when we combine two types of translation data for question alignment stage, we can harness the value of both data types and achieve higher results on non-English tasks, especially in low-resource languages.

\begingroup
\renewcommand{\arraystretch}{1.4} 
\begin{table*}[ht]
\centering
\small
\caption{Experiment results of scaling the question alignment approach to extremely large language models. For lines containing one model, the untuned model ($\mathcal{M}^-$ or $\hat{\mathcal{M}}$) is used as the base model, which is then finetuned to get the tuned version ($\mathcal{M}^+$ or $\mathcal{M}$). For lines containing three models, this means that we use small models $\mathcal{M}^+$ and $\mathcal{M}^-$ to proxy-tune the large model $\hat{\mathcal{M}}$. ``Non-En'' and ``Avg.'' denotes the average multilingual performance on non-English languages and all ten languages respectively. Bold text denotes the highest score among the same model families.}
\begin{tabular}{c|cp{2.5cm}<{\centering}p{2.5cm}<{\centering}p{2.8cm}<{\centering}ccc}
\hline
\multirow{2}{*}{\textbf{Family}} & \multirow{2}{*}{\textbf{Small tuned} $\mathcal{M}^+$} & \multirow{2}{*}{\textbf{Small untuned $\mathcal{M}^-$}}  &  \multirow{2}{*}{\textbf{Large untuned $\hat{\mathcal{M}}$}} & \multirow{2}{*}{\textbf{Large tuned $\mathcal{M}$}} & \multicolumn{3}{c}{\textbf{\textsc{mGSM}}} \\
& & & & & \textbf{Non-En} & \textbf{En} & \textbf{Avg.} \\
\hline
\multirow{8}{*}{\textit{LLaMA2}} & RAlign (7B)                    & -                   & -                & -            & 35.4   & 65.5 & 38.4 \\
& QAlign$\rightarrow$RAlign (7B) & -                   & -                & -            & 47.6   & 68.0 & 49.6 \\
& RAlign (13B)                    & -                   & -               & -            & 41.2   & 68.4 & 43.9 \\
& QAlign$\rightarrow$RAlign (13B) & -                   & -               & -            & 55.7   & 69.2 & 57.1 \\
& -                                   & -                   & -               & RAlign (70B)                    &  47.7               &  \textbf{78.4}    & 50.8 \\
& -                                   & -                   & -               & QAlign$\rightarrow$RAlign (70B) & \textbf{61.5}   & 76.0 & \textbf{63.0} \\
& QAlign$\rightarrow$RAlign (7B) & LLaMA2 (7B)         & LLaMA2 (70B)     & -            & 55.8   & 70.8 & 57.3 \\
& QAlign$\rightarrow$RAlign (13B) & LLaMA2 (13B)        & LLaMA2 (70B)    & -            & 60.1   & 76.8 & 61.8 \\
\hline
\multirow{3}{*}{\textit{LLaMA3}} & RAlign (8B)                      & -                  & -              & -             & 47.3 & 74.4 & 50.0 \\
& QAlign$\rightarrow$RAlign (8B)   & -                  & -              & -             & 58.4 & 72.0 & 59.8 \\
& QAlign$\rightarrow$RAlign (8B)   & LLaMA3 (8B)        & LLaMA3 (70B)   & -             & \textbf{64.0} & \textbf{77.2} & \textbf{65.4} \\
\hline
\multirow{4}{*}{\textit{Mistral}} & RAlign (7B)                      & -                  & -              & -                                    & 35.2 & 70.4 & 38.7 \\
& QAlign$\rightarrow$RAlign (7B)   & -                  & -              & -             & 48.2 & 70.8 & 50.4 \\
& QAlign$\rightarrow$RAlign (7B)   & Mistral (7B)       & Mixtral (8$\times$7B)  & -     & 49.4 & 74.4 & 51.9 \\
& QAlign$\rightarrow$RAlign (7B)   & Mistral (7B)       & Mixtral (8$\times$22B) & -     & \textbf{55.6} & \textbf{78.0} & \textbf{57.9} \\
\hline
\end{tabular}
\label{tab:proxy}
\end{table*}
\endgroup

\begin{figure*}[!ht]
    \centering
    \includegraphics[width=0.9\textwidth]{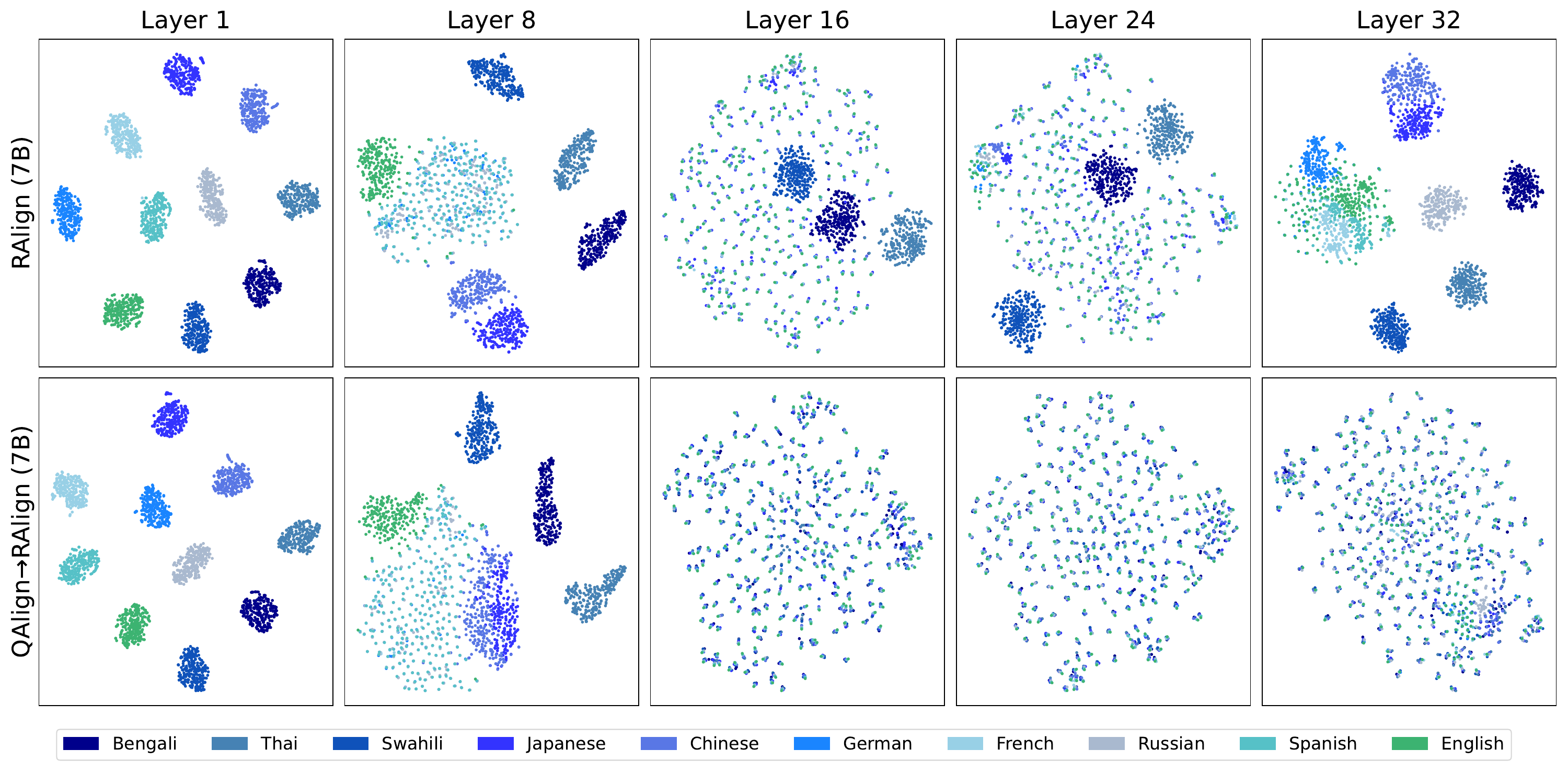}
    \caption{Visualization analysis on the representations of the last input tokens from multilingual questions. For brevity, we uniformly sample 5 layers out of the total 32 layers to illustrate the visualized distribution. Different colors denote the languages of the questions.}
    \label{fig:representation}
\end{figure*}

\subsection{The two-stage training framework scales well to extremely large language models.}
Table~\ref{tab:proxy} shows that the two-stage training framework is effective across different model sizes (7B, 8B, 13B, 70B) and scales well even on the 70B model. 
Notably, our fine-tuned LLaMA2-70B models reach a new performance ceiling on multilingual reasoning benchmarks, achieving an accuracy of 63.0\% on \textsc{mGSM}.
In the meantime, we also notice that the computational cost required for fine-tuning extremly large language models is prohibitively high\footnote{For instance, in our experiments, instruction-tuning LLaMA2-70B with \textsc{MetaMathQA} dataset (comprising 395K question-response pairs) on 8$\times$A100 GPUs takes 15 days.}.

Consequently, we explore proxy-tuning as an efficient alternative to fine-tuning.
As we introduced in Section~\ref{sec:tuning}, we use the small fine-tuned model as the expert model and its untuned version as the anti-expert model.
Across different model families (LLaMA2, LLaMA3, Mistral), including both dense models and sparse MoE models, proxy-tuning consistently enhances performance in both non-English and English tasks, effectively extrapolating our results from small models to extremely large models.
Furthermore, experiments on LLaMA2 demonstrate that carefully selecting small models for proxy-tuning, such as using LLaMA2-13B instead of using LLaMA2-7B as the small proxy model, allows us to achieve 98.1\% of the performance of fully fine-tuning (61.8 vs. 63.0) without updating any parameters in the LLaMA2-70B model.

%% file: Latex/06_analysis.tex
\section{Mechanism Analysis}
In this section, we conduct further analysis to gain a deeper understanding of the training regime and its impact on LLM's working pattern.
The analysis is performed on the instruction-tuned LLaMA2-7B model in the scenario of mathematical reasoning with chain-of-thought on \textsc{mGSM} dataset.

\subsection{Question alignment produces a more unified semantic space, facilitating the utilization of English expertise in non-English contexts}
Our analysis begins with the representation space, which fundamentally determines how the model derives the output answer.
We feed multilingual questions from the \textsc{mGSM} dataset into RAlign and QAlign$\rightarrow$RAlign models in a teacher-forcing manner, and then visualize\footnote{For visualization, we use \textsc{T-SNE} for dimension reduction.} the representations of the last input token, which decides the content of the first output tokens and serves as an important clue for how the model starts its reasoning process~\cite{wendler2024llamas}.
The visualization results are shown in Figure~\ref{fig:representation}.
For both models, the context representations of multilingual queries always stay apart in the bottom layer (1st and 8th layers).
\begingroup
\renewcommand{\arraystretch}{1} 
\begin{table*}[!ht]
\setlength{\intextsep}{0pt}
\setlength{\textfloatsep}{0pt}
\centering
\normalsize
\caption{With this case, we illustrate our computation process for evaluating reasoning path consistency.}
\scalebox{0.95}
{
\begin{tabular}{p{16cm}}
\hline
$[\textbf{Question (English)}]$ \\
There are 27 unicorns left in the world. One third of them are in the Scottish Highlands. Two thirds of the Scottish unicorns are female.  How many female Scottish unicorns are there? \\
$[\textbf{Chain-of-thought response}]$ \\
One third of the unicorns are in the Scottish Highlands, so there are 27/3 = \fbox{9} unicorns in the Highlands.\newline Two thirds of the unicorns in the Highlands are female, so there are 9 * 2/3 = \fbox{6} female unicorns in the Highlands.\newline Therefore, there are \fbox{6} female Scottish unicorns.\newline The answer is: \fbox{6} \\
$[\textbf{Extracted step-by-step answers}]$ \\
$A = [9, 6, 6, 6]$ \\
\hdashline
$[\textbf{Question (Chinese)}]$ \\
\kaiti{世界上剩下 27 只独角兽，其中三分之一在苏格兰高地。\newline 苏格兰高地的独角兽的三分之二是雌性。苏格兰雌性独角兽有多少只？} \\
$[\textbf{Chain-of-thought response}]$ \\
If there are 27 remaining unicorns in the world, and one-third of them are in the Scottish Highlands, then there are 27/3 = \fbox{9} unicorns in the Scottish Highlands.\newline If two-thirds of the unicorns in the Scottish Highlands are female, then there are 9 * 2/3 = \fbox{6} female unicorns in the Scottish Highlands.\newline Therefore, there are \fbox{6} female unicorns in the Scottish Highlands. \newline The answer is: \fbox{6} \\
$[\textbf{Extracted step-by-step answers}]$ \\
$ B = [9, 6, 6, 6]$ \\
\hline
$[\textbf{Reasoning path consistency}]$ \\
$\tilde{A} = \text{set}(A)$, $\tilde{B} = \text{set}(B)$ \\
$J(A,B)=|\tilde{A}\cap \tilde{B}|/|\tilde{A}\cup \tilde{B}| = 100\%$ \\
\hline
\end{tabular}
}
\label{tab:consistency}
\end{table*}
\endgroup
\begin{figure*}[!ht]
    \centering
    \includegraphics[width=0.9\textwidth]{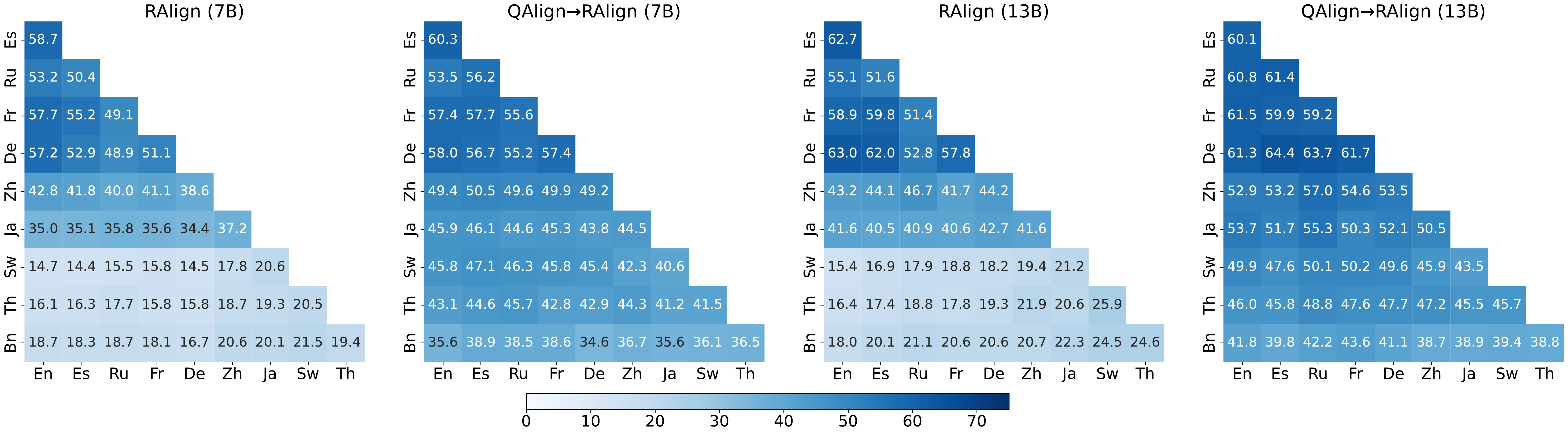}
    \caption{Comparative analysis of reasoning path consistency across different systems. The number in each cell represents the reasoning path consistency between the languages in the corresponding row and column. The number is in percentage. Darker blue denotes higher level of consistency.}
    \label{fig:consistency}
\end{figure*}
But from the middle to top layers, a notable difference emerges between our alignment-enhanced model and its unaligned counterpart: question alignment produce a more compact space, enabling the model to process multilingual queries in a unified way and facilitating the sharing of its English expertise across different languages.

\begin{figure}[!ht]
    \centering
    \includegraphics[width=0.4\textwidth]{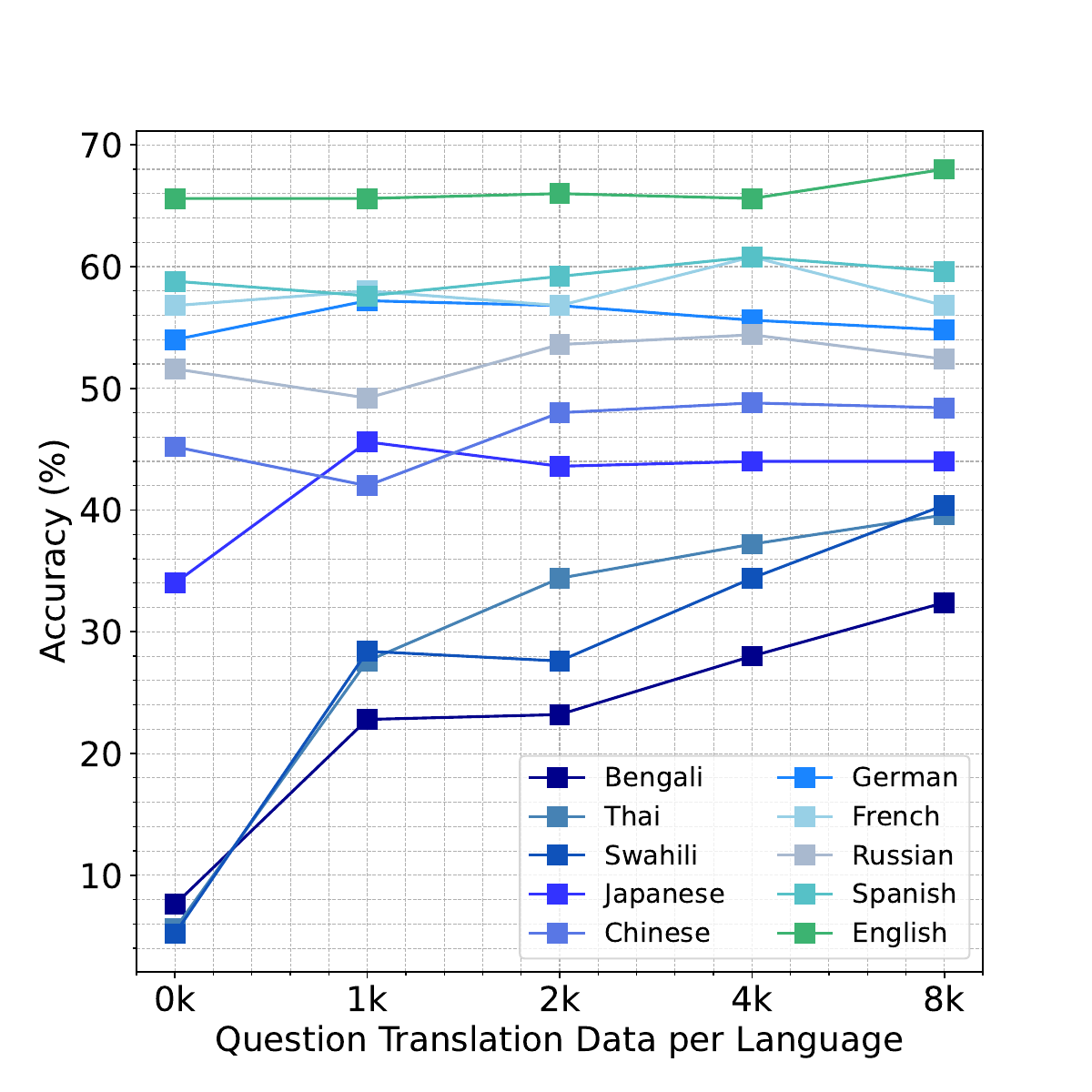}
    \caption{The impact of the size of the question translation data on multilingual reasoning accuracy.}
    \label{fig:scale}
\end{figure}
\vspace{-15pt}

\subsection{The sharing of the English expertise makes the model to employ similar problem-solving process to address multilingual questions}
In addition to analyzing the layer-wise generation process of the initial output token, we further examine step-by-step answers throughout the entire CoT output.
This analysis reveals the significant impact of the unified representation space on the reasoning process. 
We evaluate the consistency among step-by-step answers when the model is presented with the same question expressed in different languages.
Specifically, we extract intermediate computation results as lists from CoT and compute the Jaccard similarity between them to evaluate the consistency of the reasoning paths (illustrated in Table~\ref{tab:consistency})
The quantitative results are depicted in Figure~\ref{fig:consistency}.
The alignment-enhanced models (QAlign$\rightarrow$RAlign) generally have higher consistency compared to their unaligned counterparts (RAlign), particularly in low-resource languages.
This evidence further demonstrate that the question alignment approach can facilitate the sharing of reasoning abilities across languages.

\subsection{The impact of the size of question translation data}
Another important factor that influences the effectiveness of the question alignment approach is the size of the question translation data.
To analyze this factor, we uniformly downsample the multilingual question translation data from 8k pairs per language to 1k per language.
Analysis results are depicted in Figure~\ref{fig:scale}.
Generally, question alignment does not significantly affect the model's proficiency in English but does impact its performance on non-English tasks.
For high-resource languages, the performance increases gradually as the volume of translation data increases.
For low-resource languages, such as Bengali, Thai, and Swahili, scaling up the question translation data always yields substantial improvement, indicating the effectiveness and potential of this approach to enhance LLMs' performance on low-resource non-English languages.

%% file: Latex/07_conclusion.tex
\section{Conclusion}
In this paper, we build upon the pioneering work of Zhu et al. and delve deeper in multilingual reasoning to broaden the scope and provide deepened insights.
We demonstrate that the question alignment training framework can be extended to a variety of challenging reasoning scenarios and scales well to extremely large language models with either vanilla fine-tuning or efficient proxy tuning. 
In the mechanism analysis section, we provide in-depth analysis on the representation space, step-by-step answers and question translation data scales.
This analysis illuminates how the training regime influences LLM's internal representation space, thereby affecting their reasoning process. 
Additionally, the analysis on the translation data scale highlights the significant potential of the training regime to enhance the performance of LLMs on low-resource languages.

\section*{Acknowledgement}
Shujian Huang is the corresponding author. 
This work is supported by National Science Foundation of China (No. 62376116, 62176120), the Liaoning Provincial Research Foundation for Basic Research (No. 2022-KF-26-02), research project of Nanjing University-China Mobile Joint Institute.
This project has also received funding from UK Research and Innovation (UKRI) under the UK government's Horizon Europe funding guarantee (UTTER grant numbers 10039436).
Wenhao Zhu is also supported by China Scholarship Council (No.202306190172).

%% file: Latex/09_biography.tex
\section{Biography Section}
\vspace{-0.5cm}
\begin{IEEEbiography}[{\includegraphics[width=1in,height=1.25in,clip,keepaspectratio]{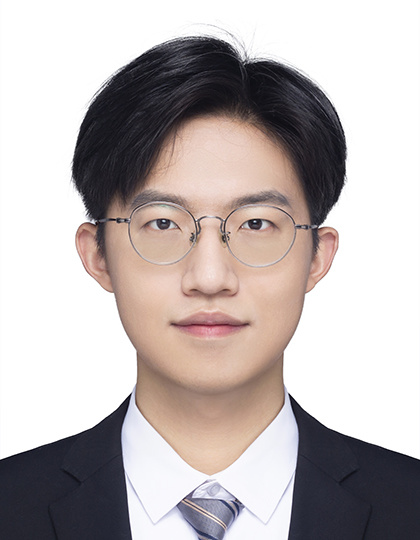}}]{Wenhao Zhu}
is a PhD student at the School of Computer Science, Nanjing University and a visiting PhD student at the Institute for Language, Cognition and Computation (ILCC), School of Informatics, University of Edinburgh.
His research focuses on multilingual large language model and machine translation. 
He has presented his work at various international conferences, including ACL, EMNLP, NAACL, NeurIPS and EACL. He also serves as a reviewer for top-tier conferences and journals, such as ACL and TPAMI.
\end{IEEEbiography}

\vspace{-0.5cm}
\begin{IEEEbiography}[{\includegraphics[width=1in,height=1.25in,clip,keepaspectratio]{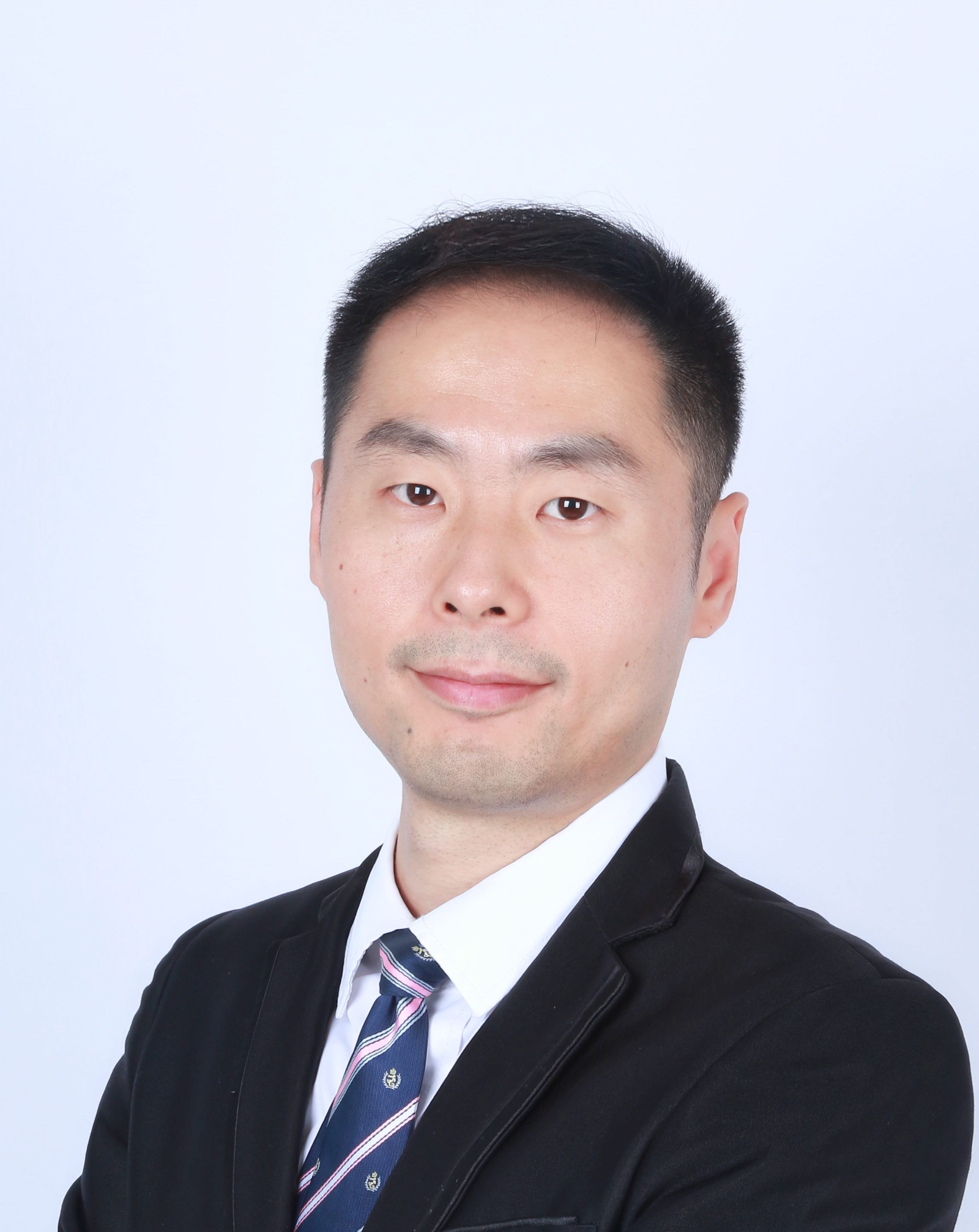}}]{Shujian Huang}
is an associate professor in School of Computer Science at Nanjing University and a member of the National Key Laboratory of Novel Software Technology. His research interests includes multilingual large language models, knowledge learning and reasoning in LLMs, etc. 
He was awarded the Excellent Young Scholar Research Project by Jiangsu Provincial Research Foundation in 2017, Outstanding Services by CIPSC in 2019, CCF-NLPCC Young Outstanding Scientist Award in 2020, CIPSC Hanwang Youth Innovation Award in 2022. 
\end{IEEEbiography}

\vspace{-0.5cm}
\begin{IEEEbiography}[{\includegraphics[width=1in,height=1.25in,clip,keepaspectratio]{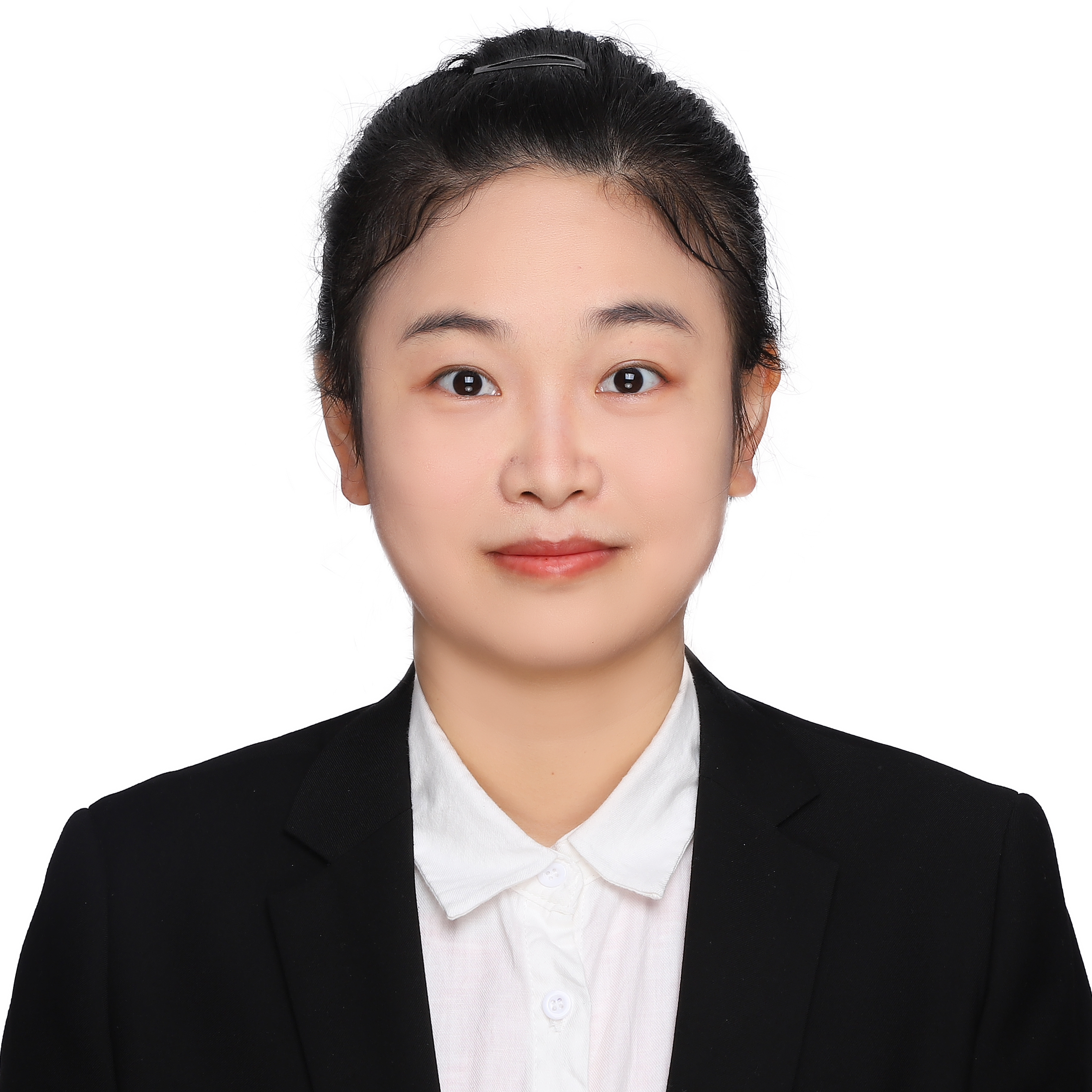}}]{Fei Yuan}
received the Ph.D degree from University of Electronic Science and Technology of China and currently serves as an researcher at the Shanghai Artificial Intelligence Laboratory. Her research interests primarily involve the development of text generation and machine learning algorithms, with a focus on advancing the multilingualism of NLP models.
\end{IEEEbiography}

\vspace{-0.5cm}
\begin{IEEEbiography}[{\includegraphics[width=1in,height=1.25in,clip,keepaspectratio]{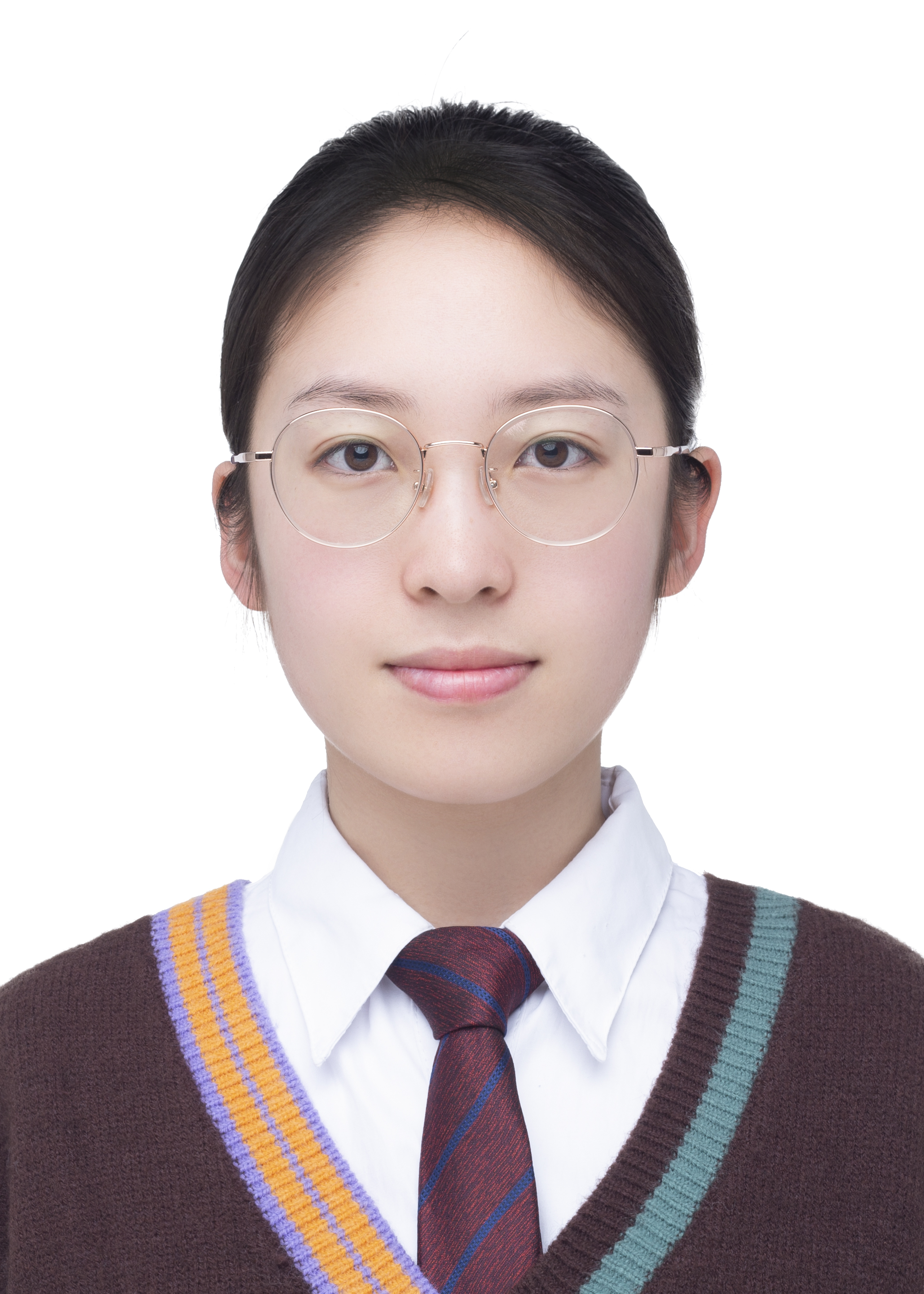}}]{Cheng Chen}
received the B.S. degree in Computer Science from Nanjing University and is currently a Master student in the Department of Computer Science and Engineering at the University of California, San Diego. Her research interests include multilingual large language models, and machine translation.
\end{IEEEbiography}

\vspace{-0.5cm}
\begin{IEEEbiography}[{\includegraphics[width=1in,height=1.25in,clip,keepaspectratio]{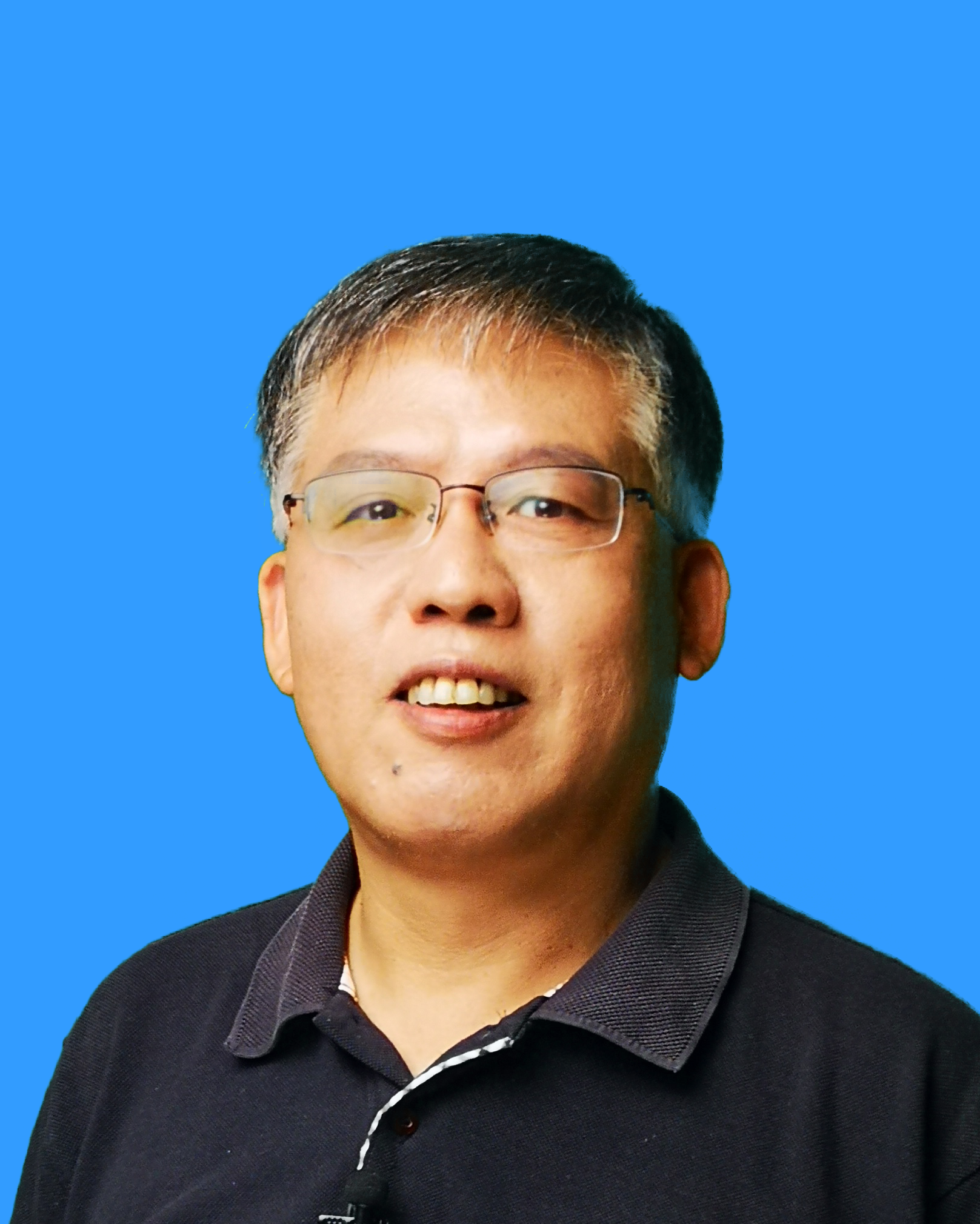}}]{Jiajun Chen}
received the B.S., M.S. and Ph.D. degrees in computer science and technology from Nanjing University. He is a Professor and Ph.D. Advisor in the School of Computer Science at Nanjing University and serves as the Director of Natural Language Processing Group. His research interest is on natural language processing, machine translation, information extraction, and text classification.
\end{IEEEbiography}

\vspace{-0.5cm}
\begin{IEEEbiography}[{\includegraphics[width=1in,height=1.25in,clip,keepaspectratio]{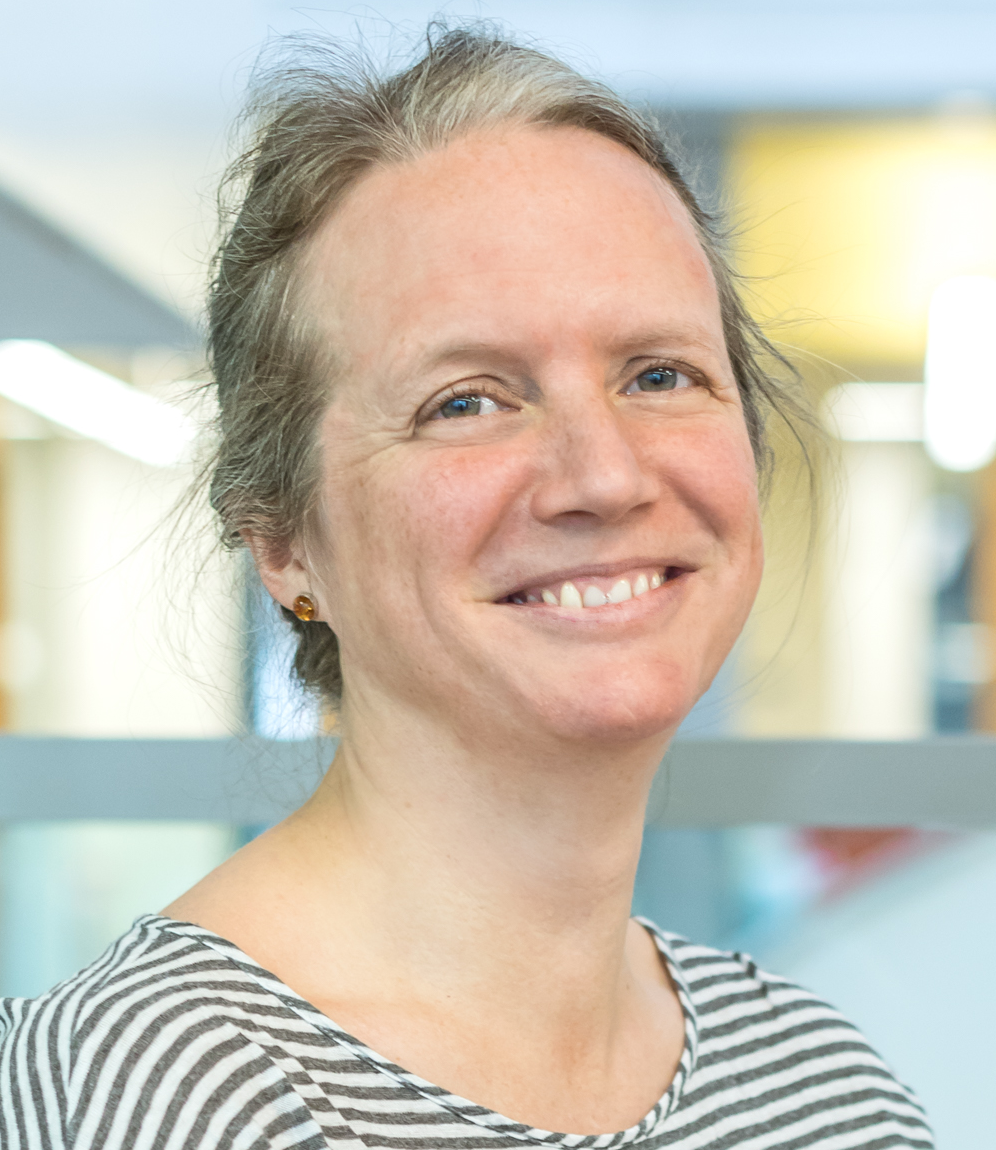}}]{Alexandra Birch}
is an Associate Professor in Natural Language Processing at the Institute for Language, Cognition and Computation (ILCC), School of Informatics, University of Edinburgh.
She is a leading member of the Statistical Machine Translation (StatMT) group and a member of the Edinburgh Natural Language Processing group (EdinburghNLP). 
Her research focuses on machine translation and multilingual dialogue. 
She has published widely in Natural Language Processing, Spoken Language, and Machine Translation venues, including ACL, EMNLP, TACL, and Computational Linguistics.
\end{IEEEbiography}

\vfill

%% file: taslp.bbl
\begin{thebibliography}{10}
\providecommand{\url}[1]{#1}
\csname url@samestyle\endcsname
\providecommand{\newblock}{\relax}
\providecommand{\bibinfo}[2]{#2}
\providecommand{\BIBentrySTDinterwordspacing}{\spaceskip=0pt\relax}
\providecommand{\BIBentryALTinterwordstretchfactor}{4}
\providecommand{\BIBentryALTinterwordspacing}{\spaceskip=\fontdimen2\font plus
\BIBentryALTinterwordstretchfactor\fontdimen3\font minus \fontdimen4\font\relax}
\providecommand{\BIBforeignlanguage}[2]{{%
\expandafter\ifx\csname l@#1\endcsname\relax
\typeout{** WARNING: IEEEtran.bst: No hyphenation pattern has been}%
\typeout{** loaded for the language `#1'. Using the pattern for}%
\typeout{** the default language instead.}%
\else
\language=\csname l@#1\endcsname
\fi
#2}}
\providecommand{\BIBdecl}{\relax}
\BIBdecl

\bibitem{hendrycks2021measuring}
D.~Hendrycks, C.~Burns, S.~Kadavath, A.~Arora, S.~Basart, E.~Tang, D.~Song, and J.~Steinhardt, ``Measuring mathematical problem solving with the {MATH} dataset,'' in \emph{Thirty-fifth Conference on Neural Information Processing Systems Datasets and Benchmarks Track (Round 2)}, 2021.

\bibitem{kojima2022large}
T.~Kojima, S.~S. Gu, M.~Reid, Y.~Matsuo, and Y.~Iwasawa, ``Large language models are zero-shot reasoners,'' in \emph{Advances in Neural Information Processing Systems (NeurIPS)}, 2022.

\bibitem{chowdhery2023palm}
A.~Chowdhery, S.~Narang, J.~Devlin, M.~Bosma, G.~Mishra, A.~Roberts, P.~Barham, H.~W. Chung, C.~Sutton, S.~Gehrmann \emph{et~al.}, ``Palm: Scaling language modeling with pathways,'' \emph{Journal of Machine Learning Research (JMLR)}, 2023.

\bibitem{touvron2023llama}
H.~Touvron, T.~Lavril, G.~Izacard, X.~Martinet, M.-A. Lachaux, T.~Lacroix, B.~Roziere, N.~Goyal, E.~Hambro, F.~Azhar \emph{et~al.}, ``Llama: Open and efficient foundation language models,'' \emph{arXiv preprint arXiv:2302.13971}, 2023.

\bibitem{shi2022language}
F.~Shi, M.~Suzgun, M.~Freitag, X.~Wang, S.~Srivats, S.~Vosoughi, H.~W. Chung, Y.~Tay, S.~Ruder, D.~Zhou \emph{et~al.}, ``Language models are multilingual chain-of-thought reasoners,'' in \emph{International Conference on Learning Representations (ICLR)}, 2022.

\bibitem{huang2023not}
H.~Huang, T.~Tang, D.~Zhang, W.~X. Zhao, T.~Song, Y.~Xia, and F.~Wei, ``Not all languages are created equal in llms: Improving multilingual capability by cross-lingual-thought prompting,'' in \emph{Findings of the Association for Computational Linguistics: EMNLP}, 2023.

\bibitem{qin2023cross}
L.~Qin, Q.~Chen, F.~Wei, S.~Huang, and W.~Che, ``Cross-lingual prompting: Improving zero-shot chain-of-thought reasoning across languages,'' in \emph{Proceedings of the Conference on Empirical Methods in Natural Language Processing (EMNLP)}, 2023.

\bibitem{blevins2022language}
T.~Blevins and L.~Zettlemoyer, ``Language contamination helps explains the cross-lingual capabilities of {E}nglish pretrained models,'' in \emph{Proceedings of the Conference on Empirical Methods in Natural Language Processing (EMNLP)}, 2022.

\bibitem{wang2023far}
Y.~Wang, H.~Ivison, P.~Dasigi, J.~Hessel, T.~Khot, K.~Chandu, D.~Wadden, K.~MacMillan, N.~A. Smith, I.~Beltagy \emph{et~al.}, ``How far can camels go? exploring the state of instruction tuning on open resources,'' \emph{Advances in Neural Information Processing Systems (NeurIPS)}, 2023.

\bibitem{nguyen2023seallms}
X.-P. Nguyen, W.~Zhang, X.~Li, M.~Aljunied, Q.~Tan, L.~Cheng, G.~Chen, Y.~Deng, S.~Yang, C.~Liu \emph{et~al.}, ``Seallms--large language models for southeast asia,'' \emph{arXiv preprint arXiv:2312.00738}, 2023.

\bibitem{chen2023breaking}
N.~Chen, Z.~Zheng, N.~Wu, L.~Shou, M.~Gong, Y.~Song, D.~Zhang, and J.~Li, ``Breaking language barriers in multilingual mathematical reasoning: Insights and observations,'' \emph{arXiv preprint arXiv:2310.20246}, 2023.

\bibitem{zhu2024question}
W.~Zhu, S.~Huang, F.~Yuan, S.~She, J.~Chen, and A.~Birch, ``Question translation training for better multilingual reasoning,'' in \emph{Findings of the Association for Computational Linguistics ACL 2024}, 2024.

\bibitem{sherborne2023optimal}
T.~Sherborne, T.~Hosking, and M.~Lapata, ``Optimal transport posterior alignment for cross-lingual semantic parsing,'' \emph{Transactions of the Association for Computational Linguistics (TACL)}, 2023.

\bibitem{wei2022chain}
J.~Wei, X.~Wang, D.~Schuurmans, M.~Bosma, b.~ichter, F.~Xia, E.~Chi, Q.~V. Le, and D.~Zhou, ``Chain-of-thought prompting elicits reasoning in large language models,'' in \emph{Advances in Neural Information Processing Systems (NeurIPS)}, 2022.

\bibitem{chen2023program}
W.~Chen, X.~Ma, X.~Wang, and W.~W. Cohen, ``Program of thoughts prompting: Disentangling computation from reasoning for numerical reasoning tasks,'' \emph{Transactions on Machine Learning Research (TMLR)}, 2023.

\bibitem{conneau2018xnli}
A.~Conneau, R.~Rinott, G.~Lample, A.~Williams, S.~R. Bowman, H.~Schwenk, and V.~Stoyanov, ``Xnli: Evaluating cross-lingual sentence representations,'' in \emph{Proceedings of the Conference on Empirical Methods in Natural Language Processing (EMNLP)}, 2018.

\bibitem{talmor2019commonsenseqa}
A.~Talmor, J.~Herzig, N.~Lourie, and J.~Berant, ``{C}ommonsense{QA}: A question answering challenge targeting commonsense knowledge,'' in \emph{Proceedings of the Conference of the North {A}merican Chapter of the Association for Computational Linguistics (NAACL)}, 2019.

\bibitem{liu2024tuning}
A.~Liu, X.~Han, Y.~Wang, Y.~Tsvetkov, Y.~Choi, and N.~A. Smith, ``Tuning language models by proxy,'' in \emph{First Conference on Language Modeling (COLM)}, 2024.

\bibitem{zhu2023multilingual}
W.~Zhu, H.~Liu, Q.~Dong, J.~Xu, S.~Huang, L.~Kong, J.~Chen, and L.~Li, ``Multilingual machine translation with large language models: Empirical results and analysis,'' in \emph{Findings of the Association for Computational Linguistics: NAACL 2024}, 2024.

\bibitem{meta2024llama3}
Meta, ``Llama3,'' \url{https://llama.meta.com/llama3/}, 2024.

\bibitem{reid2024gemini}
M.~Reid, N.~Savinov, D.~Teplyashin, D.~Lepikhin, T.~Lillicrap, J.-b. Alayrac, R.~Soricut, A.~Lazaridou, O.~Firat, J.~Schrittwieser \emph{et~al.}, ``Gemini 1.5: Unlocking multimodal understanding across millions of tokens of context,'' \emph{arXiv preprint arXiv:2403.05530}, 2024.

\bibitem{lu2024llamax}
Y.~Lu, W.~Zhu, L.~Li, Y.~Qiao, and F.~Yuan, ``Llamax: Scaling linguistic horizons of llm by enhancing translation capabilities beyond 100 languages,'' \emph{arXiv preprint arXiv:2407.05975}, 2024.

\bibitem{fujii2024continual}
K.~Fujii, T.~Nakamura, M.~Loem, H.~Iida, M.~Ohi, K.~Hattori, H.~Shota, S.~Mizuki, R.~Yokota, and N.~Okazaki, ``Continual pre-training for cross-lingual llm adaptation: Enhancing japanese language capabilities,'' \emph{arXiv preprint arXiv:2404.17790}, 2024.

\bibitem{dou2024sailor}
L.~Dou, Q.~Liu, G.~Zeng, J.~Guo, J.~Zhou, W.~Lu, and M.~Lin, ``Sailor: Open language models for south-east asia,'' \emph{arXiv preprint arXiv:2404.03608}, 2024.

\bibitem{she2024mapo}
S.~She, W.~Zou, S.~Huang, W.~Zhu, X.~Liu, X.~Geng, and J.~Chen, ``{MAPO}: Advancing multilingual reasoning through multilingual-alignment-as-preference optimization,'' in \emph{Proceedings of the 62nd Annual Meeting of the Association for Computational Linguistics (ACL)}, 2024.

\bibitem{artetxe2020translation}
M.~Artetxe, G.~Labaka, and E.~Agirre, ``Translation artifacts in cross-lingual transfer learning,'' in \emph{Proceedings of the 2020 Conference on Empirical Methods in Natural Language Processing (EMNLP)}, B.~Webber, T.~Cohn, Y.~He, and Y.~Liu, Eds., 2020.

\bibitem{yoon2024langbridge}
D.~Yoon, J.~Jang, S.~Kim, S.~Kim, S.~Shafayat, and M.~Seo, ``Langbridge: Multilingual reasoning without multilingual supervision,'' in \emph{ICLR 2024 Workshop on Mathematical and Empirical Understanding of Foundation Models}, 2024.

\bibitem{xue2021mt5}
L.~Xue, N.~Constant, A.~Roberts, M.~Kale, R.~Al-Rfou, A.~Siddhant, A.~Barua, and C.~Raffel, ``m{T}5: A massively multilingual pre-trained text-to-text transformer,'' in \emph{Proceedings of the 2021 Conference of the North American Chapter of the Association for Computational Linguistics: Human Language Technologies (NAACL-HLT)}, 2021.

\bibitem{muennighoff2023crosslingual}
N.~Muennighoff, T.~Wang, L.~Sutawika, A.~Roberts, S.~Biderman, T.~Le~Scao, M.~S. Bari, S.~Shen, Z.~X. Yong, H.~Schoelkopf, X.~Tang, D.~Radev, A.~F. Aji, K.~Almubarak, S.~Albanie, Z.~Alyafeai, A.~Webson, E.~Raff, and C.~Raffel, ``Crosslingual generalization through multitask finetuning,'' in \emph{Proceedings of the Annual Meeting of the Association for Computational Linguistics (ACL)}, 2023.

\bibitem{singh2024aya}
S.~Singh, F.~Vargus, D.~Dsouza, B.~F. Karlsson, A.~Mahendiran, W.-Y. Ko, H.~Shandilya, J.~Patel, D.~Mataciunas, L.~OMahony \emph{et~al.}, ``Aya dataset: An open-access collection for multilingual instruction tuning,'' \emph{arXiv preprint arXiv:2402.06619}, 2024.

\bibitem{pfeiffer2023modular}
J.~Pfeiffer, S.~Ruder, I.~Vuli{\'c}, and E.~Ponti, ``Modular deep learning,'' \emph{Transactions on Machine Learning Research (TMLR)}, 2023.

\bibitem{kew2023turning}
T.~Kew, F.~Schottmann, and R.~Sennrich, ``Turning english-centric llms into polyglots: How much multilinguality is needed?'' \emph{arXiv preprint arXiv:2312.12683}, 2023.

\bibitem{zhu2023extrapolating}
W.~Zhu, Y.~Lv, Q.~Dong, F.~Yuan, J.~Xu, S.~Huang, L.~Kong, J.~Chen, and L.~Li, ``Extrapolating large language models to non-english by aligning languages,'' \emph{arXiv preprint arXiv:2308.04948}, 2023.

\bibitem{zhang2023bayling}
S.~Zhang, Q.~Fang, Z.~Zhang, Z.~Ma, Y.~Zhou, L.~Huang, M.~Bu, S.~Gui, Y.~Chen, X.~Chen \emph{et~al.}, ``Bayling: Bridging cross-lingual alignment and instruction following through interactive translation for large language models,'' \emph{arXiv preprint arXiv:2306.10968}, 2023.

\bibitem{gao2023pal}
L.~Gao, A.~Madaan, S.~Zhou, U.~Alon, P.~Liu, Y.~Yang, J.~Callan, and G.~Neubig, ``Pal: Program-aided language models,'' in \emph{International Conference on Machine Learning (ICML)}.\hskip 1em plus 0.5em minus 0.4em\relax PMLR, 2023, pp. 10\,764--10\,799.

\bibitem{shi2023large}
F.~Shi, X.~Chen, K.~Misra, N.~Scales, D.~Dohan, E.~H. Chi, N.~Sch{\"a}rli, and D.~Zhou, ``Large language models can be easily distracted by irrelevant context,'' in \emph{International Conference on Machine Learning (ICML)}, 2023.

\bibitem{roziere2023code}
B.~Roziere, J.~Gehring, F.~Gloeckle, S.~Sootla, I.~Gat, X.~E. Tan, Y.~Adi, J.~Liu, T.~Remez, J.~Rapin \emph{et~al.}, ``Code llama: Open foundation models for code,'' \emph{arXiv preprint arXiv:2308.12950}, 2023.

\bibitem{jiang2024mixtral}
A.~Q. Jiang, A.~Sablayrolles, A.~Roux, A.~Mensch, B.~Savary, C.~Bamford, D.~S. Chaplot, D.~de~las Casas, E.~B. Hanna, F.~Bressand, G.~Lengyel, G.~Bour, G.~Lample, L.~R. Lavaud, L.~Saulnier, M.-A. Lachaux, P.~Stock, S.~Subramanian, S.~Yang, S.~Antoniak, T.~L. Scao, T.~Gervet, T.~Lavril, T.~Wang, T.~Lacroix, and W.~E. Sayed, ``Mixtral of experts,'' \emph{arXiv preprint arXiv:2401.04088}, 2024.

\bibitem{mistral2024mixtral}
Mistral, ``https://mistral.ai/news/mixtral-8x22b,'' 2024.

\bibitem{yuan2023scaling}
Z.~Yuan, H.~Yuan, C.~Li, G.~Dong, C.~Tan, and C.~Zhou, ``Scaling relationship on learning mathematical reasoning with large language models,'' \emph{arXiv preprint arXiv:2308.01825}, 2023.

\bibitem{toshniwal2024openmathinstruct}
S.~Toshniwal, I.~Moshkov, S.~Narenthiran, D.~Gitman, F.~Jia, and I.~Gitman, ``Openmathinstruct-1: A 1.8 million math instruction tuning dataset,'' \emph{arXiv preprint arXiv:2402.10176}, 2024.

\bibitem{lin2021xcsr}
B.~Y. Lin, S.~Lee, X.~Qiao, and X.~Ren, ``Common sense beyond english: Evaluating and improving multilingual language models for commonsense reasoning,'' in \emph{Proceedings of the Annual Meeting of the Association for Computational Linguistics (ACL)}, 2021.

\bibitem{williams2018broad}
A.~Williams, N.~Nangia, and S.~Bowman, ``A broad-coverage challenge corpus for sentence understanding through inference,'' in \emph{Proceedings of the 2018 Conference of the North {A}merican Chapter of the Association for Computational Linguistics: Human Language Technologies (NAACL-HLT)}, 2018.

\bibitem{schwenk2021wikimatrix}
H.~Schwenk, V.~Chaudhary, S.~Sun, H.~Gong, and F.~Guzm{\'a}n, ``{WikiMatrix: Mining 135M Parallel Sentences in 1620 Language Pairs from Wikipedia},'' in \emph{Proceedings of Conference of the European Chapter of the Association for Computational Linguistics (EACL)}, 2021.

\bibitem{yu2023metamath}
L.~Yu, W.~Jiang, H.~Shi, J.~Yu, Z.~Liu, Y.~Zhang, J.~T. Kwok, Z.~Li, A.~Weller, and W.~Liu, ``Metamath: Bootstrap your own mathematical questions for large language models,'' \emph{arXiv preprint arXiv:2309.12284}, 2023.

\bibitem{zhao2021neurst}
C.~Zhao, M.~Wang, Q.~Dong, R.~Ye, and L.~Li, ``{NeurST}: Neural speech translation toolkit,'' in \emph{the Annual Meeting of the Association for Computational Linguistics (ACL)}, 2021.

\bibitem{luo2023wizardmath}
H.~Luo, Q.~Sun, C.~Xu, P.~Zhao, J.~Lou, C.~Tao, X.~Geng, Q.~Lin, S.~Chen, and D.~Zhang, ``Wizardmath: Empowering mathematical reasoning for large language models via reinforced evol-instruct,'' \emph{arXiv preprint arXiv:2308.09583}, 2023.

\bibitem{yue2023mammoth}
X.~Yue, X.~Qu, G.~Zhang, Y.~Fu, W.~Huang, H.~Sun, Y.~Su, and W.~Chen, ``Mammoth: Building math generalist models through hybrid instruction tuning,'' \emph{arXiv preprint arXiv:2309.05653}, 2023.

\bibitem{ivison2023camels}
H.~Ivison, Y.~Wang, V.~Pyatkin, N.~Lambert, M.~Peters, P.~Dasigi, J.~Jang, D.~Wadden, N.~A. Smith, I.~Beltagy, and H.~Hajishirzi, ``Camels in a changing climate: Enhancing lm adaptation with tulu 2,'' 2023.

\bibitem{burchell2023open}
L.~Burchell, A.~Birch, N.~Bogoychev, and K.~Heafield, ``An open dataset and model for language identification,'' in \emph{Proceedings of the Annual Meeting of the Association for Computational Linguistics (ACL)}, 2023.

\bibitem{wendler2024llamas}
C.~Wendler, V.~Veselovsky, G.~Monea, and R.~West, ``Do llamas work in {E}nglish? on the latent language of multilingual transformers,'' in \emph{Proceedings of the 62nd Annual Meeting of the Association for Computational Linguistics (ACL)}, 2024.

\end{thebibliography}
